\documentclass[acmlarge]{acmart}

\AtBeginDocument{%
  }

\setcopyright{none}
\makeatletter
\renewcommand{\@copyrightowner}{}
\renewcommand{\@copyrightpermission}{}
\renewcommand{\@formatdoi}[1]{}
\renewcommand{\@copyrightyear}{}
\makeatother

\makeatletter
\def\@acmReferenceFormat#1{}
\def\@acmPages{}
\makeatother

\usepackage{algorithm}
\usepackage{algpseudocode}
\usepackage{multirow}
\usepackage{graphicx}
\usepackage{subcaption}
\usepackage{xcolor}
\usepackage{caption}
\usepackage[toc]{appendix}
\usepackage[normalem]{ulem}

\setcopyright{none}
\makeatletter
\renewcommand{\@copyrightowner}{}
\renewcommand{\@copyrightpermission}{}
\renewcommand{\@formatdoi}[1]{}
\renewcommand{\@copyrightyear}{}
\makeatother


\usepackage[normalem]{ulem}
\usepackage[dvipsnames]{xcolor} 

\usepackage[toc]{appendix}  

\makeatletter
\def\@acmReferenceFormat#1{} 
\def\@acmPages{} 
\makeatother

\begin{document}

\title{FT-ARM: Fine-Tuned Agentic Reflection Multimodal Language Model for Pressure Ulcer Severity Classification with Reasoning}

\author{Reza Saadati Fard}
\email{rsaadatifard@wpi.edu}
\authornote{Corresponding author.}

\author{Emmanuel Agu}
\email{emmanuel@wpi.edu}

\author{Palawat Busaranuvong}
\email{pbusaranuvong@wpi.edu}

\author{Deepak Kumar}
\email{dkumar1@wpi.edu}

\author{Shefalika Gautam}
\email{sgautam@wpi.edu}

\author{Bengisu Tulu}
\email{bengisu@wpi.edu}

\author{Diane Strong}
\email{dstrong@wpi.edu}

\author{Lorraine Loretz}
\email{loretzdpmnp@gmail.com}

\affiliation{%
  \institution{Worcester Polytechnic Institute}
  \city{Worcester}
  \state{MA}
  \country{USA}
}

\affiliation{%
  \institution{UMass Memorial Health}
  \city{Worcester}
  \state{MA}
  \country{USA}
}

\renewcommand{\shortauthors}{Saadati Fard et al.}

\begin{abstract}
Pressure Ulcers (PUs) are a prevalent and serious healthcare concern. To guide appropriate treatment, accurate categorization of PU severity into one of four categories (I - IV) is required. However, as severity categories often have  subtle and subjective visual distinctions, manual staging is challenging and prone to variability across clinicians, necessitating automated solutions. Prior AI-driven approaches explored Convolutional Neural Networks (CNN) and Vision Transformers (ViT), which yielded promising image classification results but their predictions had limited interpretability. Multimodal large language models (MLLMs), which integrate vision and language understanding, are an emerging paradigm for contextualized and explainable image classification. We present FT-ARM (Fine-Tuned Agentic Reflection Multimodal model), which combines a fine-tuned MLLM with an agentic self-reflection mechanism to classify pressure ulcer image severity classification, while also providing rich rationale (reasoning) and context for its classifications. Inspired by diagnostic reassessments by human clinicians,  FT-ARM's self-reflection strategy, performs  iterative self-refinement of its initial predictions by reasoning over visual features and  encoded clinical knowledge (via natural language understanding of clinical notes) to improve classification accuracy and consistency. 
In experiments on the publicly available Pressure Injury Image Dataset (PIID), our fine-tuned model—FT-ARM with LLaMA 3.2 90B as backbone—achieved an accuracy of 85\% in classifying pressure ulcer stages I–IV, outperforming prior CNN-based models (by +4\%). It is also instructive to note that prior work utilizing CNN or ViT models, typically reported model performance in offline evaluations, which would likely degrade in live deployments. In contrast, FT-ARM is designed for and evaluated in a live inference scenario that reflect real-time deployment conditions, enhancing its potential for clinical application. Beyond predictive performance, FT-ARM generates clinically grounded natural language explanations (reasons) for each prediction, offering interpretability aligned with expert reasoning. By combining fine-tuning with reflective reasoning on multimodal inputs, FT-ARM advances the reliability, transparency, and clinical utility of automated wound assessment systems—addressing a critical need for consistent and explainable pressure ulcer staging to support improved patient care.
\end{abstract}



\settopmatter{printacmref=false}
\maketitle

\section{Introduction}

\textbf{Motivation} Pressure ulcers (PUs) – also known as pressure injuries – are a significant healthcare issue, especially in immobilized or elderly patients. PUs affect about 12.8\% of hospitalized adults, presenting a substantial burden on healthcare systems \cite{lei2025convolutional, aldughayfiq2023yolo}. PUs can cause severe pain and complications; in advanced stages they may lead to serious infections or even death, underscoring the importance of early and accurate identification, and treatment \cite{aldughayfiq2023yolo}.

\begin{figure}[h]
\centering
\includegraphics[width=0.7\linewidth]{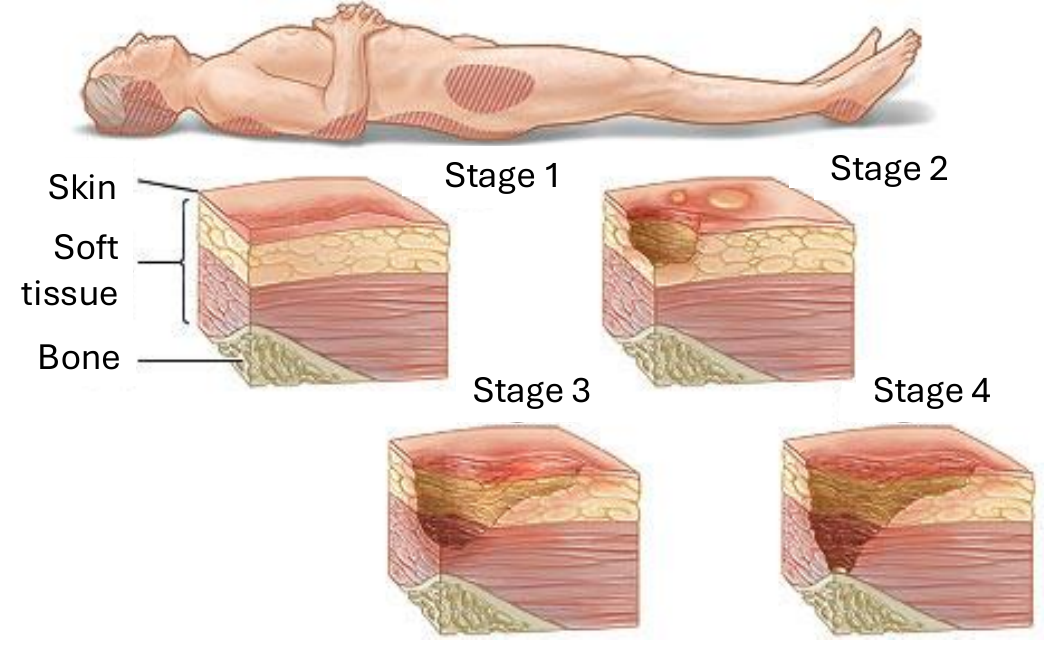}
\caption{Visualization of the anatomy of pressure ulcer stages I–IV, as defined by  the National Pressure injury Advisory Panel (NPIAP)~\cite{edsberg2016revised}. Stage I involves skin erythema without tissue loss; Stage II presents partial-thickness skin loss with exposure of dermis; Stage III shows full-thickness tissue loss extending into subcutaneous fat; and Stage IV indicates extensive damage reaching a muscle or bone.~\cite{barghouthi2023systematic}.}
\label{fig:piid-stages}
\end{figure}

\textbf{Background and challenges} Accurate staging is crucial for selecting appropriate treatment~\cite{lei2025convolutional}. The National Pressure Injury Advisory Panel (NPIAP) classifies pressure injuries into four severity stages ~\cite{edsberg2016revised}. An overview of these stages is shown in Figure~\ref{fig:piid-stages}, which illustrates the increasing severity and depth of tissue damage from stage I to IV. However, as distinctions between pressure ulcer stages are often subtle, determining the correct stage via visual inspection (e.g. at the point of care) is often challenging.
Moreover, wound lighting, angle, or skin tone are additional, well-documented  challenges for visual pressure ulcer staging~\cite{ay2022deep}.

Manual classification of pressure ulcer stages from its visual appearance is inherently subjective. Wound assessments by clinicians suffer from considerable inter-rater variability \cite{lei2025convolutional}, and accuracy often depends on the clinician’s training and experience. Studies report that in practice, only 23–58\% of medical staff are able to correctly classify pressure ulcers~\cite{lei2025convolutional}. Less experienced providers, in particular, struggle with staging consistency~\cite{aldughayfiq2023yolo}.  When staging is incorrect, diagnosis and treatment may be wrong or delayed, potentially worsening the injury, increasing the risk of complications such as infection or necrosis, and ultimately leading to longer hospital stays and higher medical costs \cite{wang2024novel}.

These challenges highlight the need for a more objective and reliable automated method for PU staging~\cite{aldughayfiq2023yolo} that can improve consistency and support patient management. Researchers have noted that AI-driven clinical decision systems are more likely to be trusted when they include reasoning components—particularly in high-stakes domains such as wound care~\cite{tonekaboni2019clinicians, ribeiro2016should}. Self-explaining models that generate not only decisions but also interpretable rationales can foster clinician confidence, reduce uncertainty, and help correct diagnostic errors made by the AI~\cite{sonoda2025structured, busaranuvong2025explainable}. Recent advances in medical AI have demonstrated the effectiveness of multimodal learning for clinical tasks~\cite{fard2025multimodal, brehmer2025fine}, further motivating the neeed for explainable multimodal AI-driven methods for PU staging that in addition to classifying ulcer stages, also provides  clinically meaningful justifications (reasons).

\textbf{Prior work:} Researchers have recently explored computer vision and deep learning approaches to automate the severity classification of PU images. Convolutional neural network (CNN) models, which learn visual features of wounds, have demonstrated promising results~\cite{lei2025convolutional, aldughayfiq2023yolo}. For example, a deep CNN (DenseNet121) trained on pressure injury photographs achieved approximately 93.7\% accuracy in staging~\cite{lei2025convolutional}. Vision Transformer (ViT)-based models have also been applied to various medical classification tasks~\cite{busaranuvong2024guided, cho2025development}; one ViT-based approach reported an accuracy of approximately 97.8\% for PU staging~\cite{cho2025development}. These results demonstrate the feasibility of accurate, automated staging using image-only models.

Despite their success, vision-only classifiers have notable limitations. They typically require large labeled image datasets for training and are constrained to analyzing only visual features of wounds~\cite{ferber2024context}. Crucial contextual information—such as clinical notes, patient-specific factors, clinical variables such as those contained in the Electronic Health Records (EHRs), or stage guidelines and formal definitions—are not incorporated into image-only approaches. Moreover, such classifiers often lack interpretability and typically do not review or revise their initial decisions iteratively to improve performance on ambiguous or challenging PU cases.

\textbf{Multimodal Large Language Models (MLLMs)} These limitations of image classification prompted interest in more intelligent AI systems that combine image recognition with higher-level reasoning. In the broader AI field, the emergence of Multimodal Large Language Models (MLLMs) that can perform visual as well as textual analyses  (such as GPT-4 Vision) offers a new approach for medical image and data analysis \cite{li2023systematic}. MLLMs integrate visual analyses with the language-based knowledge and reasoning of LLMs, enabling functionality such as answering questions about an image or generating a descriptive reports~\cite{li2023systematic, rezaei2025agentic}. Notably, a recent study demonstrated that given only a few example prompts, a GPT-4 Vision-based model could match or even outperform task-specific neural networks on certain medical image classification tasks~\cite{ferber2024context}. This finding suggests that foundation models that are already trained on massive datasets can be adapted to specialized tasks using in-context learning on minimal additional data to leverage their vast pre-trained knowledge. However, general-purpose LLMs such as GPT-4o are not optimized for domain-specific medical tasks out-of-the-box. They require adaptation—via fine-tuning or structured prompting—to perform reliably on medical tasks. Moreover, closed-source models such as GPT-4o do not support optimization via fine-tuning, limiting their ability to learn domain- and task-specific visual and linguistic information. Prior work demonstrated that clinician trust in medical AI improves significantly with diagnostic transparency when Large Language Models (LLMs) provide structured reasoning—such as Chain-of-Thought prompting or decision rationale~\cite{sonoda2025structured, busaranuvong2025explainable}. This is especially valuable in complex tasks such as wound classification, where interpretability is critical.

\textbf{Our approach} We propose  Fine-Tuned Agentic Reflection Multimodal model (FT-ARM) for PU severity classification. FT-ARM adapts a state-of-the-art MLLM to the pressure wound staging task via domain-specific fine-tuning. FT-ARM accepts multimodal input consisting of a wound photograph and a standardized prompt that asks the model to classify the stage of the pressure ulcer. While the system can accept additional optional textual input—such as brief clinician notes—for enhanced reasoning, all reported results in this study were generated using only the prompt without supplemental clinical context. – enabling it to reason about the wound's appearance in light of clinical knowledge such as definitions of each stage’s characteristics. A key innovation of our approach is the incorporation of an agentic reflection mechanism. Inspired by cognitive processes and recent AI agentic frameworks, FT-ARM does not make only a single forward prediction. Instead, first, it produces an initial classification with associated explanatory rationale. Additional passes are then made to review and refine its output iteratively by “reflecting” on potential errors and improving performance on areas of uncertainty. Specifically, FT-ARM's architecture incorporates a Generator LLM that produces an initial diagnosis and rationale, followed by a Critique LLM that analyzes the response and suggests revisions based on criteria in clinical guidelines. While the ability to reference clinical guidelines (e.g. via Retrieval Augmented Generation (RAG)) is powerful, we defer exploring that direction to future work. Specifically, in this project, the critique LLM does not utilize any information from clinical guidelines. Instead, it simply double-checks classification results via multiple reflection passes to ensure that they are consistent. The generator then incorporates feedback from the critique LLM  in a refinement loop, to finalize its prediction. This reflection process empowers FT-ARM to detect and correct initially erroneous decisions, much like a clinician double-checks an initial diagnosis to ensure correctness. Prior works have demonstrated that such self-reflection can significantly improve the problem-solving accuracy of LLM-based agents~\cite{madaan2023self, renze2024self, rafi2024enhancing, gou2023critic}. Additionally, CNN- and ViT-based image analyses models are typically evaluated in offline settings using static test datasets; however, their performance often degrades in real-world deployment due to variations in imaging conditions, and clinical environments~\cite{oakden2020hidden}. In contrast, FT-ARM is explicitly designed and evaluated during inference in live deployment. Consequently, its reported accuracy more reliably reflects real-world performance and avoids the accuracy drops commonly experienced in models and approaches developed and evaluated offline. In summary, FT-ARM marries advanced image recognition with language-driven reasoning to address the critical challenges of pressure ulcer staging. Tuning (FT) enables FT-ARM to learn domain-specific visual cues of each ulcer stage, while its Agentic Reflection (AR) strategy adds a layer of interpretative verification to its predictions. To the best of our knowledge, our work is the first to employ an MLLM with an internal agentic reflection reasoning loop for wound care. 

\begin{figure}[h]
\centering
\includegraphics[width=0.9\linewidth]{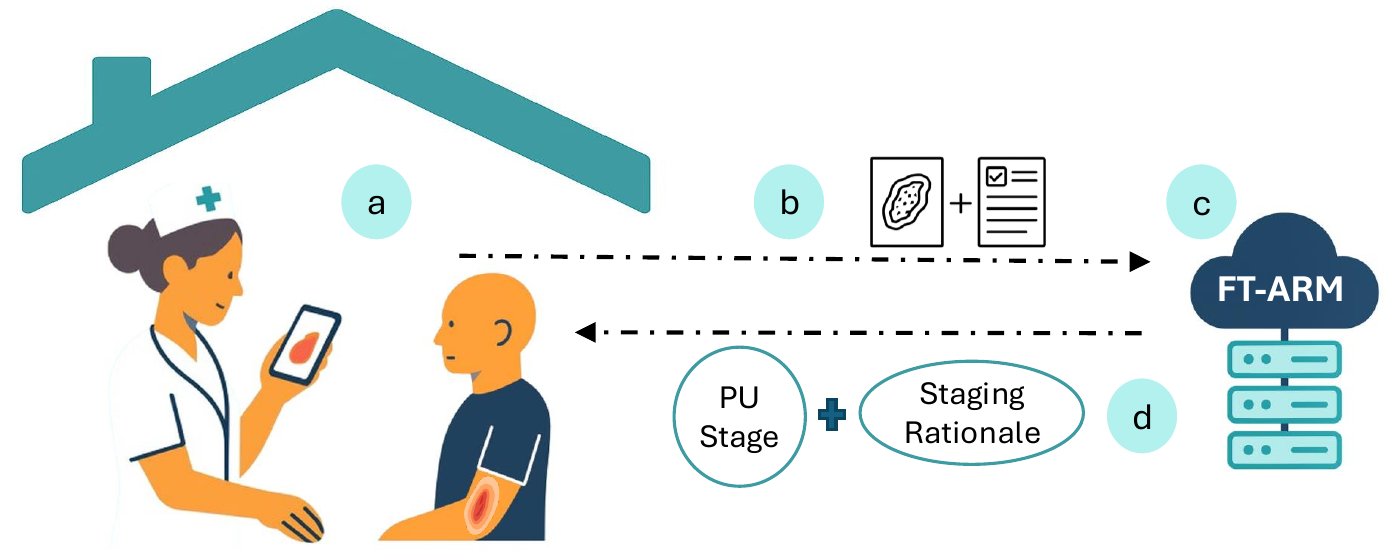}
\caption{Envisioned usage scenario for FT-ARM. A nurse captures a wound photo and optional clinical notes using a smartphone app (a), which sends them to FT-ARM running in the cloud (b–c), and receives a predicted pressure ulcer stage with decision rationale and explanations (d).}
\label{fig:envisioned-scenario}
\end{figure}

\textbf{Envisioned usage scenario} Figure~\ref{fig:envisioned-scenario} presents an envisioned deployment scenario of FT-ARM. A nurse captures a pressure wound image and possibly enters clinical notes into a smartphone wound app during a visit to a wound patient at their home. The wound image and optional clinical notes are sent to FT-ARM running in the cloud. FT-ARM analyzes the inputs and returns a predicted PU ulcer stage along with a clinically valid rationale, enabling real-time, interpretable decision support in point-of-care settings. This illustration highlights FT-ARM configured as a black-box service—accepting multimodal input and returning structured outputs.

FT-ARM was rigorously evaluated on the Pressure Injury Image Dataset (PIID)~\cite{ay2022deep}, which consists of 1,091 smartphone-captured stages I–IV pressure ulcer images with ground-truth annotations by trained experts. In rigorous evaluation, FT-ARM outperformed a comprehensive set of state-of-the-art (SOTA) CNN and ViT image-only classification baselines. These baselines included MobileNetV2 \cite{sandler2018mobilenetv2}, VGG16 \cite{simonyan2015vgg}, DenseNet121 \cite{huang2017densenet}, and ResNet152 \cite{xie2017resnext} from Ay et al.~\cite{ay2022deep}, as well as EfficientNetV2-s \cite{tan2021efficientnetv2}, ResNeXt50~\cite{xie2017resnext} with weighted Feature Pyramid Network (wFPN) \cite{li2019weighted}, and the Swin Transformer-tiny ViT model by Wang et al.~\cite{wang2024novel}. FT-ARM achieved 85\% pressure stage classification accuracy on a PIID test set, outperforming the previous SOTA CNN-based model—ResNeXt50 with a weighted Feature Pyramid Network—which achieved 81.5\% accuracy~\cite{wang2024novel}, and also surpassed Swin Transformer-tiny \cite{liu2021swin}, the top ViT baseline that achieved 75.5\% accuracy.

Beyond accuracy, FT-ARM outputs not only the stage of the pressure ulcer in an input image but also a clinically-valid explanation of the corresponding rationale in natural language. By aligning its justifications with clinical reasoning, FT-ARM offers interpretable outputs that foster confidence and support usability in real-world deployment. By providing fast, consistent, and explainable pressure ulcer assessments, FT-ARM has the potential to enhance decision support in wound care and improve patient outcomes. FT-ARM also builds on pre-trained MLLMs and adapts them efficiently using lightweight fine-tuning (LoRA) and self-reflection, significantly reducing the data size and computational resources required. This enables FT-ARM to be deployed after tuning on a single, modestly sized dataset (such as PIID), making it far more suitable for fast adaptation and clinical translation in real-world medical settings where annotated data is scarce and deployment timelines are constrained.

\textbf{Contributions} of this paper are:
\begin{itemize}
    \item We introduce FT-ARM, a fine-tuned agentic reflection multimodal model for pressure ulcer (PU) staging from images, capable of performing both visual analysis and clinically grounded language reasoning.
    \item We design a two-stage reflection mechanism, consisting of Generator and Critique LLMs, that iteratively reassess outputs in a manner analogous to human diagnostic review, thereby improving classification accuracy.
    \item We apply parameter-efficient Low-Rank Adaptation (LoRA) fine-tuning to adapt the base MLLM for PU staging, enabling effective training with limited data and compute resources.
    \item We conduct rigorous evaluation on the publicly available PIID dataset, showing that FT-ARM achieves 85.2\% accuracy and 0.85 $F_1$-score, surpassing strong CNN and ViT baselines (best accuracy 81.5\%).
    \item The explanatory decision rationale output by FT-ARM were validated via review by am experienced wound care nurse, who interpreted them and confirmed that they are largely clinically meaningful. 
\end{itemize}

The remainder of this paper is structured as follows: Section~\ref{sec:background} presents background on pressure ulcer staging, MLLMs, and output refinement strategies. Section~\ref{sec:relatedwork} reviews related work on wound classification using CNNs, ViTs, and multimodal approaches. Section~\ref{sec:methodology} describes the FT-ARM architecture, including the fine-tuning process and agentic reflection mechanism. Section~\ref{sec:evaluation} details the experimental setup, baseline models, and performance results. Section~\ref{sec:discussion} analyzes key findings, limitations, and implications for clinical deployment. Finally, Section~\ref{sec:conclusion} concludes the paper and outlines future work.

\section{Background}
\label{sec:background}
\subsection{Pressure Ulcer Staging and the PIID Dataset}

Ay et al.\cite{ay2022deep} introduced the Pressure Injury Image Dataset (PIID), a public dataset consisting of smartphone wound images in RGB format along with expert annotations of pressure ulcer stages I to IV as defined by the European Pressure Ulcer Advisory Panel (EPUAP) guidelines. Each wound image is $299\times299$ pixels in resolution. Images were collected from patients aged 50–88 at Erzurum Regional Training and Research Hospital, between 2018 and 2021, using smartphone cameras under varied lighting and imaging conditions. To address class imbalance, 15 additional Stage I images were sourced via Google Images~\cite{ay2022deep}. All images were labeled by medical doctors according to EPUAP staging criteria~\cite{edsberg2016revised}. The PIID has diverse patient demographics, wound location, and skin tones, providing a realistic benchmark for machine learning model development. Figure~\ref{fig:piid-stages} shows representative images from each PIID class/stage, and Table~\ref{tab:piid-dist} summarizes the class distribution. To the best of our knowledge, PIID remains the only publicly available PU image dataset annotated by medical professionals, containing wounds in all four stages. While a few other PU datasets exist (e.g., Kaggle~\cite{kaggle2023ulcers}), their labels were not annotated by experts, they did not follow established clinical guidelines for pressure ulcer staging or their labeling procedures were inconsistent, limiting their utility in clinically-valid research.

\begin{table}[h]
\centering
\caption{Distribution of stages of the pressure ulcer images in the PIID dataset.}
\label{tab:piid-dist}
\begin{tabular}{|c|c|}
\hline
\textbf{Stage} & \textbf{Number of Images} \\
\hline
Stage I & 230 \\
Stage II & 313 \\
Stage III & 275 \\
Stage IV & 273 \\
\hline
\end{tabular}
\end{table}

\subsection{Multimodal Large Language Models (MLLMs)}

Multimodal large language models (MLLMs) are sophisticated AI models capable of analyzing both textual and non-textual inputs such as images, and producing classifications as well as coherent natural language outputs, image-grounded answers, textual explanations, and reasoning chains. MLLMs integrate vision and language understanding in a unified architecture, enabling contextual reasoning across modalities~\cite{openai2024gpt4o}. 

Unlike traditional LLMs such as the original Generative Pre-trained Transformer (GPT)~\cite{brown2020gpt3} that analyzes only text inputs, MLLMs also analyze images by incorporating a vision encoder—often a convolutional network or vision transformer—to transform images into embeddings that the Language Model (LM) can reason over. MLLMs that support both image and text modalities are typically trained on large datasets of image–text pairs enabling them to learn cross-modal information and reason across modalities. State of the art examples of MLLMs include OpenAI’s GPT-4o~\cite{openai2024gpt4o}, Meta’s LLaMA 3.2~\cite{meta2024llama3}, Mistral AI’s Pixtral-12B~\cite{mistral2023mixtral}, Alibaba’s Qwen-VL~\cite{alibaba2024qwen2}, and DeepSeek VL2~\cite{deepseek2024vl2}. These models jointly analyze visual and textual inputs to generate fluent, context-aware natural language responses grounded in image content. By combining visual recognition with language-based reasoning, MLLMs are able to support tasks such as image captioning, visual question answering, and medical image interpretation~\cite{li2023systematic}.

In clinical applications such as PU staging, MLLMs can assess wound images while leveraging medical domain knowledge to guide classification decisions. When available, they can also analyze and incorporate accompanying clinical notes provided by wound experts, enabling deeper contextual understanding and improved diagnostic accuracy in complex scenarios. This multimodal reasoning capability is notably absent in traditional deep learning models such as CNN- and ViT-based classifiers, which solely analyze visual input and cannot incorporate textual context, or provide clinical reasoning.

\subsection{Fine-Tuning of Pre-trained Models}


Fine-tuning is the process of using a smaller, domain-specific dataset to adapt a pre-trained model to a specific task ~\cite{parthasarathy2024ultimate}. Several fine-tuning strategies have been proposed especially to adapt deep neural networks, and mostly differ in which and how many parameters are updated. \textit{Full fine-tuning} updates all or most of a model’s parameters~\cite{parthasarathy2024ultimate}. While effective for small models,  full fine-tuning is generally unsuitable for large language models due to high computational cost, risk of overfitting on limited data, and potential degradation of pre-trained knowledge~\cite{xu2023parameter}. \textit{Partial fine-tuning} improves on efficiency by updating only a subset of parameters—typically the last layers—while keeping the rest frozen~\cite{parthasarathy2024ultimate}. However, partial fine-tuning still alters the core model weights and yields limited savings on the number of model parameters. ~\textit{Adapter-based tuning}, a form of Parameter-Efficient Fine-Tuning (PEFT), introduces small trainable modules (e.g., adapters or low-rank matrices) while keeping the original model weights frozen~\cite{xu2023parameter}. This results in efficient task adaptation with minimal resource overhead.

Given the large scale of the backbone MLLMs it utilizes—ranging from several billion to over a trillion parameters—FT-ARM adopts an adapter-based fine-tuning approach. Full and partial fine-tuning were impractical due to their high resource demands and the risk of degrading general-purpose prediction capabilities. Specifically, FT-ARM uses Low-Rank Adaptation (LoRA)~\cite{hu2022lora}, a PEFT method that introduces small trainable matrices into selected layers of the MLLM —typically within the attention projections—while keeping all base model weights frozen. This design enables lightweight and scalable adaptation with minimal computational overhead. Unlike conventional CNN or ViT fine-tuning that often require re-training the entire model on large image datasets, LoRA allows efficient adaptation of large multimodal LLMs. It is particularly well-suited for domain-specific medical tasks such as PU classification, where labeled data is limited, expensive to collect,  and where preserving the model's pre-trained general knowledge is essential.

\subsection{LLM Output Refinement}

LLM output refinement refers to strategies to improve the quality, accuracy and reliability of responses generated, either through self-evaluation or external feedback~\cite{brinkmann2025self}. A range of refinement approaches exist. Reinforcement Learning from Human Feedback (RLHF)~\cite{ouyang2022training} relies on expert-labeled data and model retraining, which is resource-intensive and unsuitable for real-time applications. In contrast, self-reflective methods—such as Self-Refine~\cite{madaan2023self} and Critic-in-the-loop~\cite{gou2023critic}—enable models to evaluate and iteratively revise their outputs without model re-training. These techniques generally involve generating an initial response, identifying mistakes, and refining the result in a feedback loop. FT-ARM performs output refinement via an internal self-refinement mechanism involving two LLMs: a Generator produces an initial stage prediction and rationale, and a Critic evaluates and provides feedback. If inconsistencies are found, the Generator updates its output accordingly. This iterative loop improves accuracy and consistency, making it suitable for clinical tasks such as PU classification.

\section{Related Work}
\label{sec:relatedwork}
Since its public release,  only a small number of studies have utilized the PIID for pressure injury  machine learning analyses or  algorithm development. To the best of our knowledge, just two prior works—Ay et al.~\cite{ay2022deep} and Wang et al.~\cite{wang2024novel}—have performed comprehensive machine learning modeling utilizing the  PIID. However, they explored image-only machine learning methods that analyzed only PIID images using CNNs or ViTs, with offline evaluation on static train-test dataset splits. In our evaluation, these works serve as baselines against which FT-ARM is compared.

\textbf{CNN-Based Methods on PIID:} Ay et al.~\cite{ay2022deep} benchmarked six CNN architectures on the PIID dataset utilizing data augmentation and transfer learning. CNN architectures included DenseNet121 \cite{huang2017densenet}, ResNet50/152 \cite{xie2017resnext}, InceptionV3 \cite{szegedy2016inceptionv3}, MobileNetV2 \cite{sandler2018mobilenetv2}, and VGG16 \cite{simonyan2015vgg}. DenseNet121 achieved the highest overall accuracy of approximately 75\%, with per-class performance ranging from 55\% to 77\%, underscoring the difficulty of distinguishing visually similar stages (e.g., Stage III vs. Stage IV). Building on this work, Wang et al.~\cite{wang2024novel} extended the PIID with 1,519 additional clinical images, and evaluated deeper CNNs such as EfficientNetV2 \cite{tan2021efficientnetv2} and ResNeXt50 \cite{xie2017resnext}. Their best-performing model—ResNeXt50 with a weighted Feature Pyramid Network (wFPN) \cite{li2019weighted}—achieved a new SOTA accuracy of 81.5\%, with $F_1$-score of 0.811, precision of 0.808, and recall of 0.816. Notably, the accuracy of classifying 
Stage III pressure ulcers, one of the most error prone, improved from 60\% to 76\%.

\textbf{ViT-Based Methods on PIID:} In addition to CNNs, Wang et al.~\cite{wang2024novel} explored ViT-based approaches for PU severity classification on the PIID.  Swin Transformer-tiny~\cite{liu2021swin} model, the best-performing ViT-based model, achieved 75.5\% classification accuracy, which was comparable to earlier CNN baselines. Although it performed slightly below the wFPN-enhanced CNN, its competitive performance highlighted the promise of transformer architectures for PU staging and serves as motivation for further exploration of multimodal transformer models for this task. Table~\ref{tab:piid-comparison} summarizes the  results reported by all published machine learning studies that utilized the PIID.

\begin{table}[h]
\centering
\caption{Comparison of published methods using the PIID dataset for pressure ulcer stage classification.}
\label{tab:piid-comparison}
\begin{tabular}{|l|l|l|c|}
\hline
\textbf{Authors} & \textbf{Method} & \textbf{Model Type} & \textbf{Accuracy} \\
\hline
Ay et al., 2022 \cite{ay2022deep}& MobileNetV2 & CNN & 54.84\% \\
Ay et al., 2022 \cite{ay2022deep}& VGG16 & CNN & 71.89\% \\
Ay et al., 2022 \cite{ay2022deep}& DenseNet121 & CNN & 67.28\% \\
Ay et al., 2022 \cite{ay2022deep}& ResNet152 & CNN & 77.42\% \\
Wang et al., 2024 \cite{wang2024novel}& EfficientNetV2-s & CNN & 78.8\% \\
Wang et al., 2024 \cite{wang2024novel}& ResNeXt50 & CNN & 79.5\% \\
Wang et al., 2024 \cite{wang2024novel}& ResNeXt50 + wFPN & CNN & \textbf{81.5\%} \\
Wang et al., 2024 \cite{wang2024novel}& Swin Transformer-tiny & ViT & 75.5\% \\
\hline
\end{tabular}
\end{table}

\begin{table}[h]
\centering
\caption{Summary of prior machine learning work that did not utilize the PIID dataset.}
\label{tab:nonpiid-comparison}
\begin{tabular}{|p{3.2cm}|p{3.5cm}|p{1.8cm}|p{4.0cm}|p{1.5cm}|}
\hline
\textbf{Authors} & \textbf{Method} & \textbf{Model Type} & \textbf{Dataset (size, source, \#classes)} & \textbf{Accuracy} \\
\hline
Kosmopoulos and Tzevelekou, 2007 \cite{kosmopoulos2007automated} & SVM on image segment features & Classic ML & 85 images, private, 4 classes & $\sim$80\% \\
Veredas et al., 2015 \cite{veredas2015wound} & SVM, RF, neural network & Classic ML & 113 images, private, 4 classes & 87.77\% \\
Seo et al., 2023 \cite{seo2023visual} & VGG16, ResNet-50/152, EfficientNet-B4 & CNN & 2461 images, private , 4 classes & 91.46\% \\
Lau et al., 2022 \cite{lau2022artificial} & YOLOv4 + data augmentation & CNN & 190 images, Medetec (public), 3 classes & 73.3\% \\
Xu et al., 2022 \cite{xu2022classification} & Modified DeIT & ViT & 2918 images, private (DFU), 3 classes & 78.0\% \\
Brehmer et al., 2025 \cite{brehmer2025fine} & TinyViT + multimodal fusion & ViT & 763 images (PU only), private, 4 stages & 82.55\% \\
Cho and Yoo, 2025 \cite{cho2025development} & PUC-ViT with MixUp + semi-supervised training & ViT & 395 images, private, 5 stages & 95.6\% \\
\hline
\end{tabular}
\end{table}

\textbf{Pressure Ulcer Non-PIID Studies:} While Ay et al.~\cite{ay2022deep} and Wang et al.~\cite{wang2024novel} provide the most comprehensive evaluations till date, other researchers have explored machine learning approaches using other datasets, summarized in Table~\ref{tab:nonpiid-comparison}. Prior work on non-PIID PU datasets, were often tailored to specific clinical needs or considered constrained use cases. Due to limited dataset availability, class imbalance, and differences in target labels, the results achieved by these studies are not directly comparable to those that analyzed the PIID. Instead such wide variability that limit direct comparison underscores the importance of the PIID as a standardized benchmark for developing reproducible and generalizable machine learning pressure ulcer staging models.
Kosmopoulos and Tzevelekou~\cite{kosmopoulos2007automated} used SVM on image segment features to classify 85 private pressure ulcer images. Veredas et al.~\cite{veredas2015wound} applied SVM, Random Forest, and neural networks to 113 images, achieving good performance. Their best SVM-based model achieved 87.77\% accuracy on wound-bed classification, though the dataset was relatively small and their results are not directly comparable to those achieved on the PIID dataset. Seo et al.~\cite{seo2023visual} achieved 91.46\% accuracy using EfficientNet-B4 on 2461 private images for nursing-focused wound classification. While the performance achieved is impressive, the dataset is not publicly available.

Lau et al.~\cite{lau2022artificial} employed YOLOv4 on 190 public images from the Medetec dataset. Their model classified three classes with an accuracy of 73.3\% in real-time but utilized a limited dataset. Xu et al.~\cite{xu2022classification} used a ViT-based model on 2918 diabetic foot ulcer images, classifying three classes with an accuracy of 78.0\%. Brehmer et al.~\cite{brehmer2025fine} developed a multimodal TinyViT model for PU staging using 763 images and achieved 82.55\% accuracy across four stages. Cho and Yoo~\cite{cho2025development} trained a ViT-based PUC-ViT model on 395 PU images covering five stages, reporting 97.76\% accuracy. In spite of the strong performance, the dataset was small and private, increasing concerns about generalizability.

\section{Methodology}
\label{sec:methodology}

\subsection{FT-ARM Architecture and Multimodal Integration}

Our proposed \textbf{FT-ARM} model adopts a unified multimodal architecture that directly incorporates image and text analysis into a single LLM. Unlike prior approaches that rely on external vision transformers~\cite{dosovitskiy2020image}, FT-ARM integrates an MLLM backbone that simultaneously analyzes visual and textual inputs. This reflects the current shift in MLLM design toward decoder-only models that fuse modalities that are jointly analyzed within a single computation graph~\cite{yin2024survey,zhang2025unified}.

\begin{figure}[h]
    \centering
    \includegraphics[width=0.95\textwidth]{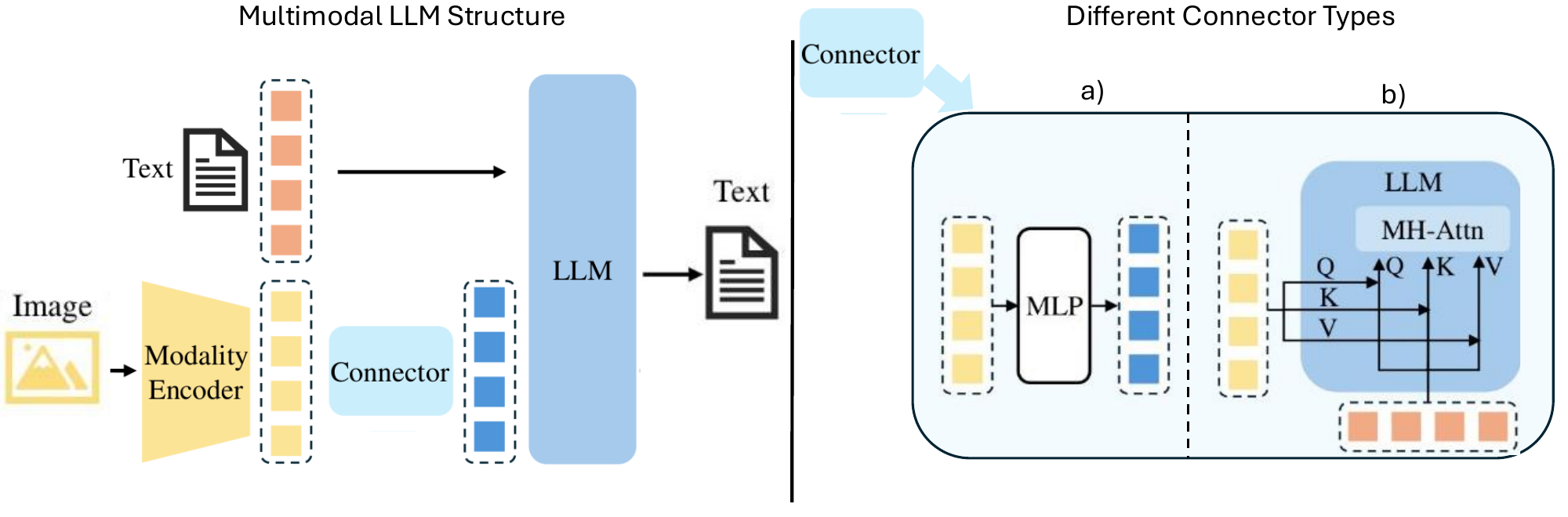}
    \caption{Overview of a typical Multimodal LLM (MLLM) architecture \cite{yin2024survey}.. The left side shows a standard processing pipeline in which text and image inputs are embedded via a tokenizer and modality encoder, respectively. A connector module then aligns the image embeddings with text tokens before feeding them into a unified LLM for response generation. The right side illustrates two common connector types: \textbf{(a)} Projection-based connectors (e.g., MLPs), which transform visual embeddings into token space; and \textbf{(b)} Fusion-based connectors, which integrate image features directly within the LLM via multi-head attention.}
    \label{fig:mllm-architecture}
\end{figure}

An input wound image, a textual prompt (e.g., asking the MLLM to determine the PU stage), and optional caregiver-provided clinical notes (e.g., brief metadata or wound descriptions) are jointly analyzed by the MLLM using a visual encoder and a language tokenizer. The image is encoded into visual embeddings, while the text is tokenized into language-compatible tokens. These modality-specific representations are then aligned into a shared embedding space via either projection-based or fusion-based connectors (See Figure \ref{fig:mllm-architecture}). Projection-based connectors—used in FT-ARM’s base model (LLaMA 3.2)—employ Multi-Layer Perceptrons (MLPs)to map visual features into the LLM’s token space~\cite{mu2025mmxu}, enabling unified token-level reasoning. Similar to how they are utilized in GPT-4o~\cite{openai2024gpt4o}, fusion-based connectors, integrate image and text features directly within the transformer layers, allowing deeper multimodal cross-modal learning~\cite{yin2024survey}.

Figure~\ref{fig:mllm-architecture} illustrates the architecture of MLLMs. On the left, the general structure of MLLMs is shown: image and text inputs are processed independently—images via a modality encoder and text via tokenization—before being fused via a connector module. The fused embeddings are then passed to the LLM for downstream reasoning.
On the right, two types of connector strategies are compared:
(a) Projection-based connectors, which use MLPs or similar modules to transform image embeddings into language-compatible vectors that are then concatenated with text tokens before being fed into the LLM. This approach is used in MLLMs such as LLaMA 3.2~\cite{meta2024llama3}, Pixtral \cite{mistral2023mixtral}, Qwen-VL 2.5 \cite{alibaba2024qwen2}, and DeepSeek-VL \cite{deepseek2024vl2}.
(b) Fusion-based connectors, which enable direct cross-modal attention within the LLM. Here, image and text embeddings interact at every layer via multi-head attention mechanisms, as seen in GPT-4o \cite{openai2024gpt4o}.

\begin{figure}[h]
\centering
\includegraphics[width=0.95\linewidth]{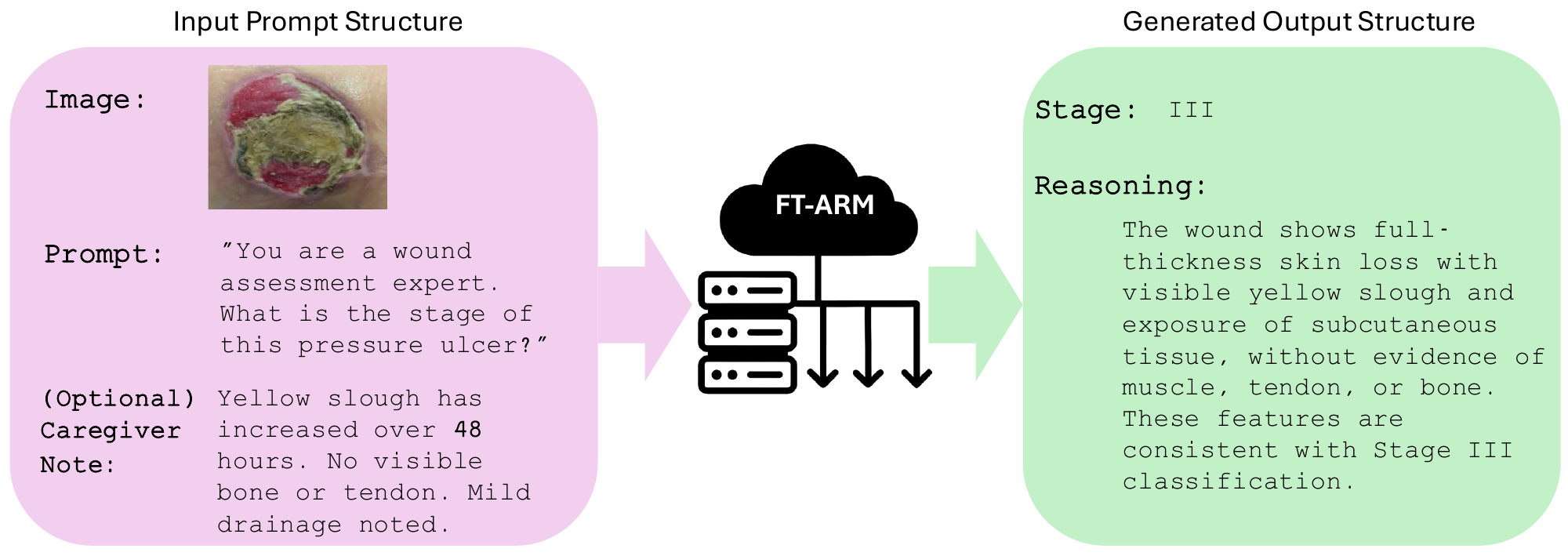}
\caption{Example of FT-ARM input and output structure. The input consists of a wound image, a task-specific prompt, and an optional caregiver note. While caregiver notes are supported by the system, they were not used during fine-tuning or evaluation in this study. FT-ARM generates a structured output that includes both the predicted PU stage and corresponding explanatory rationale. This example illustrates a Stage III prediction from visual and contextual features.}
\label{fig:prompt-output-structure}
\end{figure}

FT-ARM’s LLMs are trained and evaluated on the pressure ulcer staging task, where it receives a prompt such as: \emph{“You are a wound  expert. Based on analyses of the image and context, determine the stage of this pressure ulcer.”}. We adopt an instruction-style prompting strategy similar to the "task prefixing" method often used in instruction tuning~\cite{wei2021finetuned}. In our setup, each input consists of a wound image and a task-specific prompt such as "What is the stage of this pressure ulcer?", and optionally, a structured clinical note. While our current training and evaluation pipeline does not include clinical notes, our proposed system is designed to accept them in future if caregivers provide additional textual descriptions of the wound. Figure~\ref{fig:prompt-output-structure} shows an example of the complete input-output structure used by FT-ARM, including the image, prompt, optional caregiver note, and the stage classification with associated rationale generated.

\begin{figure}[h]
\centering
\includegraphics[width=0.8\linewidth]{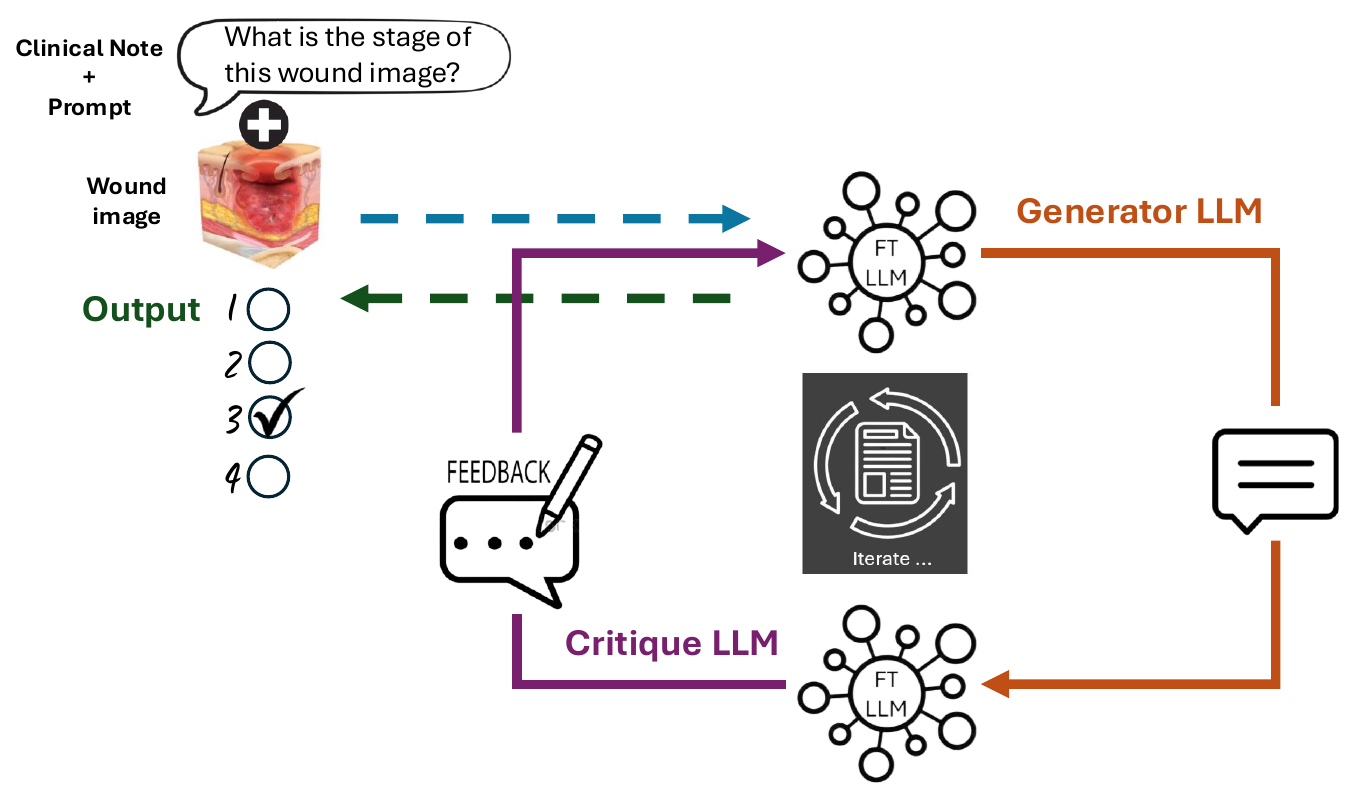}
\caption{FT-ARM architecture for PU staging. A wound image and optional clinical note are fed into a fine-tuned Generator LLM, which generates an initial stage prediction and corresponding explanatory rationale. This output is then reviewed by a Critique LLM that provides feedback through a self-reflection loop, enabling the system to revise its answer. The final output includes both a pressure ulcer stage classification and an interpretable corresponding rationale. This iterative structure enhances both predictive reliability and clinical transparency.}
\label{fig:ftarm-arch}
\end{figure}

FT-ARM is trained on the PIID dataset to perform pressure ulcer staging as a generative language modeling task, where the model outputs both a classification label and  clinical rationale.  This dual-output design is an important characteristic of FT-ARM, combining label prediction with interpretability. By leveraging its pre-trained multimodal capabilities and domain-specific fine-tuning, FT-ARM's MLLMs learn to identify key visual indicators of severity (e.g., tissue loss, slough, or exposed structures) and to contextualize them in clinical language. This design supports high performance while maintaining interpretability—two attributes that are important in medical decision support systems.
Figure~\ref{fig:ftarm-arch} summarizes the complete FT-ARM architecture. A wound image and prompt are embedded and passed to a fine-tuned Generator LLM to produce a stage prediction and rationale (see Section~\ref{sec:lora}). This output is reviewed by a Critique LLM through an iterative reflection loop (detailed in Section~\ref{sec:arm}), enabling prediction refinement.

\subsection{Multimodal LLM Backbone}

FT-ARM is designed to be modular and can employ various alternate MLLMs as its backbone. This flexibility will enable FT-ARM to leverage and benefit from improvements in the evolving MLLM landscape, making it adaptable to newer, more capable models as they become available in future. In future, more powerful  MLLMs with improved vision–language alignment or clinical grounding can be integrated into the FT-ARM framework without redesigning its architecture.
To evaluate currently available options, we experimented with SOTA MLLMs, including GPT-4o \cite{openai2024gpt4o}, LLaMA 3.2 \cite{meta2024llama3}, DeepSeek-VL \cite{deepseek2024vl2}, Qwen~2.5 \cite{alibaba2024qwen2}, and Pixtral-12B \cite{mistral2023mixtral}. Table~\ref{tab:mllm-properties} summarizes their key attributes and limitations. While all support image input and text output, they differ in model size, fine-tuning capability, and multi-image input support—a factor that influenced few-shot prompting experiments in our evaluation.

\begin{table}[h]
\centering
\caption{Comparison of MLLMs evaluated for use as backbones in FT-ARM.}
\label{tab:mllm-properties}
\begin{tabular}{|l|l|c|c|c|}
\hline
\textbf{Model} & \textbf{Developer} & \textbf{Params} & \textbf{Fine-Tunable?} & \textbf{Multi-Image Input} \\
\hline
GPT-4o & OpenAI &  - & No & Yes \\
GPT-4o-mini & OpenAI & - & No & Yes \\
LLaMA 3.2 & Meta & 11B / 90B & Yes & No \\
Pixtral-12B & Mistral AI & 12B & Yes & No \\
Qwen2-VL-72B & Alibaba & 72B & Yes & Yes \\
DeepSeek-VL & DeepSeek & \textasciitilde{}13B & No & Yes \\
\hline
\end{tabular}
\end{table}

Only MLLMs that have made their source code publicly available such as LLaMA, Pixtral, and Qwen2-VL can support fine-tuning via LoRA, which is utilized to develop FT-ARM. In contrast, closed source models such as GPT-4o, GPT-4o-mini, and DeepSeek-VL do not support fine-tuning for domain-specific adaptation. While GPT-4o permits some limited fine-tuning to support general-purpose language tasks via its API, it does not permit fine-tuning for specialized medical applications such as those involving clinical images. Additionally, at the time of writing, DeepSeek-VL has not released documentation or tools for performing fine-tuning on their multimodal models, making it unsuitable for task-specific adaptation. Furthermore, models such as LLaMA and Pixtral are not able to accept multiple input images. These architectural constraints precluded our use of these MLLMs and directly influenced our backbone MLLM selection and evaluation strategy.

\subsection{Agentic Reflection Mechanism (ARM)}
\label{sec:arm}

Agentic Reflection Mechanism (ARM) is a key innovation of FT-ARM's. ARM is an iterative self-critique loop inspired by recent self-reflection strategies in AI agents \cite{renze2024self}. A two-stage generator–critic cycle enables FT-ARM to evaluate and refine its own predictions. This mechanism addresses the ambiguity of certain PU cases by encouraging the model to “think twice” about the output.

\textbf{Stage 1: Generative Reasoning (Generator).} In the first stage, the LLM (acting as the generator) analyzes the image and produces an initial classification along with a rationale. For instance, given the embedded image features, the model might initially output: “The wound likely corresponds to Stage III, as there is full-thickness skin loss and visible fat tissue.” This response is generated in free-form text, which includes both a tentative pressure ulcer stage prediction and the model’s reasoning (rationale). 

\textbf{Stage 2: Self-Critique (Critic).}

In the second stage, FT-ARM employs a  Critique LLM to review the Generator’s output. While in principle the Generator and Critique LLMs can be instantiated as different models, in this project, the same underlying MLLM is utilized for both roles, switching prompts to elicit different behaviors. This separation of the generator and critique LLMs encourages independent reasoning and reduces the risk of self-reinforcing errors \cite{renze2024self}. The Generator’s initial output is passed back to the Critique LLM using a structured critique prompt that asks the model to assess the accuracy, clinical plausibility, and completeness of the original prediction and associated rationale. The Critique LLM reviews the reasoning, checks the visual features, and suggests corrections if inconsistencies, errors, or oversights are detected. Although it does not access formal staging guidelines in this paper, the Critique LLM performs this assessment based on its internal reasoning and visual evidence alone.

\begin{figure}[t]
\centering
\includegraphics[width=0.9\textwidth]{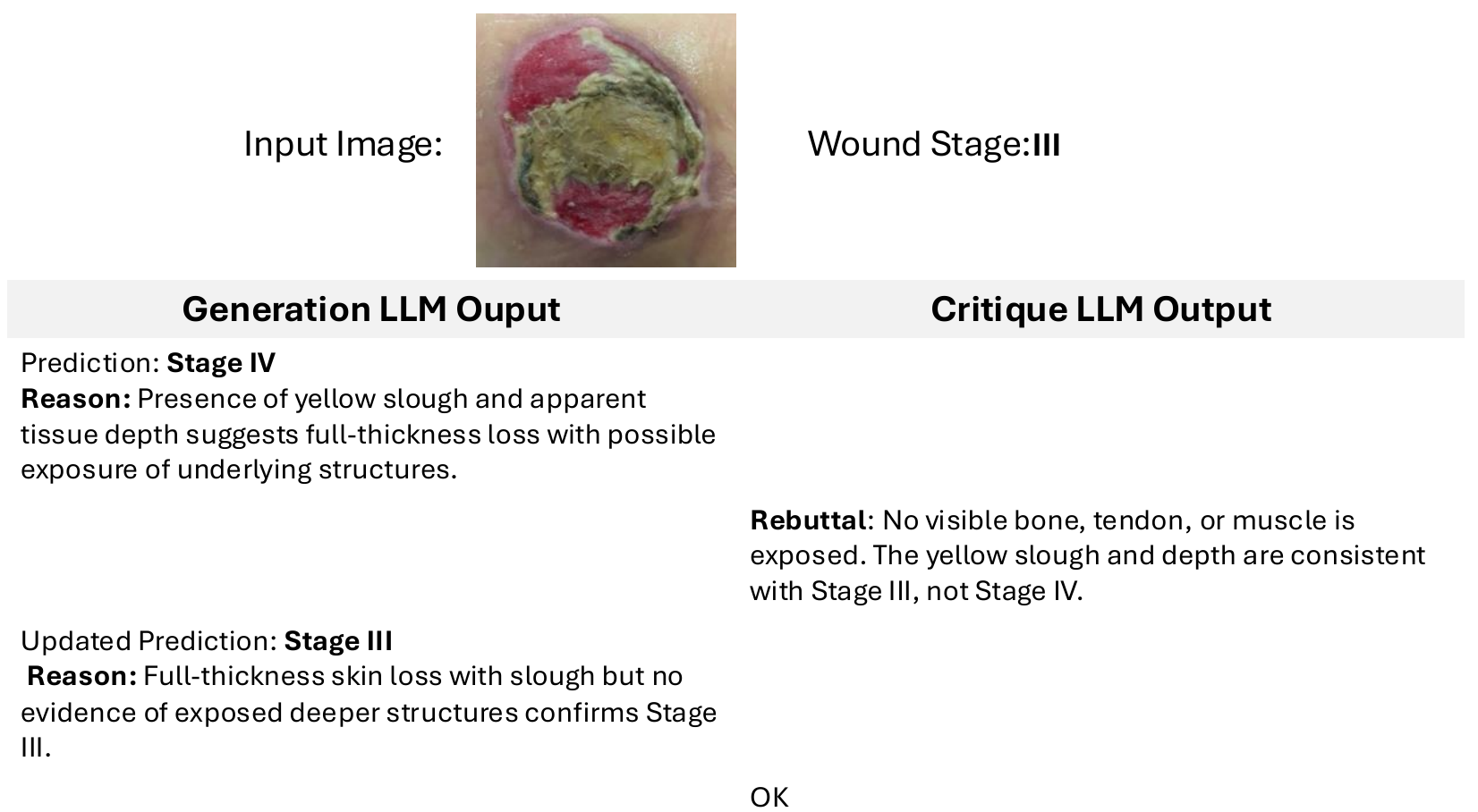}
\caption{Example of the FT-ARM agentic reflection mechanism. Based on the presence of slough and wound depth, the Generator LLM initially predicts that the pressure ulcer is in stage IV. The Critique LLM identifies that deeper structures such as bone and muscle are not visible and proposes Stage III as being more appropriate. The Generator then updates its answer and rationale, aligning with guidelines. This step-wise revision enables FT-ARM to resolve ambiguous cases, and improve diagnostic precision.}
\label{fig:agentic-example}
\end{figure}

Figure~\ref{fig:agentic-example} illustrates this process: the Generator initially predicts Stage IV based on tissue depth and slough, while the Critique LLM identifies that no bone or tendon is visible and recommends Stage III. The Generator then updates its answer accordingly. This rebuttal-feedback loop enables FT-ARM to improve the quality of its decisions by revisiting and potentially revising its initial judgment.

\textbf{Refinement Loop:} The feedback from the critic is then used to refine the pressure ulcer stage classification. In our implementation, one full generator–critic cycle (i.e., two stages) is performed, which yields a final answer after the critique. The generator re-evaluates its initial decision in light of the critic’s comments and generates a revised output. This final output is the PU stage classification reported. In the example, the model might revise the answer to: “Final verdict: This ulcer is Stage IV, as there is necrosis and possible bone exposure.”  The full procedure is detailed in Algorithm~\ref{alg:ft-arm}.

\begin{algorithm}[t]
\caption{Agentic Reflection Mechanism (ARM) for MLLMs}
\label{alg:ft-arm}
\begin{algorithmic}[1]
\State \textbf{Input:} Image–text prompt set, Generation Prompt $P_g$, Critique Prompt $P_c$, max iterations $N=2$
\For{each image and prompt}
    \State Use $P_g$ to generate initial classification and rationale (Generator)
    \State Save as Initial Response
    \For{$i = 1$ to $N$}
        \State Feed current response into $P_c$ (Critic)
        \State Critic evaluates reasoning, visual cues, and plausibility
        \If{Critic outputs ``OK''}
            \State \textbf{break}
        \Else
            \State Generate critique feedback
            \State Inject feedback into $P_g$
            \State Generator re-evaluates and produces revised prediction
        \EndIf
    \EndFor
    \State Finalize the prediction
\EndFor
\State Calculate final classification accuracy after ARM
\end{algorithmic}
\end{algorithm}

We found that up to two generator–critic iterations were effective in improving accuracy and rationale quality. However, more iterations yielded diminishing returns. For example, in well-defined wound cases such as Stage I ulcers, further iterations sometimes caused the model to second-guess its already correct classifications, introducing uncertainty such as revising its initial Stage I classificaiton to Stage II. In more ambiguous cases, additional iterations often produced more verbose outputs without improving accuracy. Therefore, based on these empirical findings, we set a two-iteration threshold in FT-ARM, unless the Generator explicitly accepts the critique by outputting “OK,” which terminates the loop early.

Figure~\ref{fig:agentic-example} presents an example of this mechanism in action. The model initially predicts that the pressure ulcer is in Stage IV due to the presence of slough and wound depth. The Critique LLM identifies that deeper structures such as bone or muscle are not visible and suggests that the wound is more consistent with Stage III. The model then updates its classification accordingly, refining both the label and its rationale. In this way, the agentic reflection loop reduces classification errors and improves the alignment of FT-ARM's predictions with established clinical guidelines.

\subsection{Fine-Tuning with LoRA}
\label{sec:lora}

\subsubsection{Low Rank Adaptation}
To enable PU classification and explanation, FT-ARM is fine-tuned as a generative multimodal model. Given an input image $I$ and optional textual context $T$, the model is trained to output a sequence $Y = (y_1, y_2, \dots, y_L)$ consisting of both the predicted PU stage and supporting rationale. Fine-tuning is supervised using a standard autoregressive language modeling loss function:

\begin{equation}
\mathcal{L}_{\text{LM}}(\theta) = -\sum_{t=1}^{L} \log P_\theta(y_t^* \mid y_{<t}^*, I, T),
\end{equation}

where $(y_1^*, \dots, y_L^*)$ is the ground-truth output sequence. This loss encourages the model to generate both the correct classification and a clinically coherent explanation. Although the stage prediction task can be viewed as classification over a discrete set of labels $s$, this framework naturally embeds classification within the token-level generation process, where the probability of the true stage $s^*$ is captured by $P_\theta(s^* \mid I, T)$ as part of the full sequence output.

\begin{figure}[h]
\centering
\includegraphics[width=0.7\linewidth]{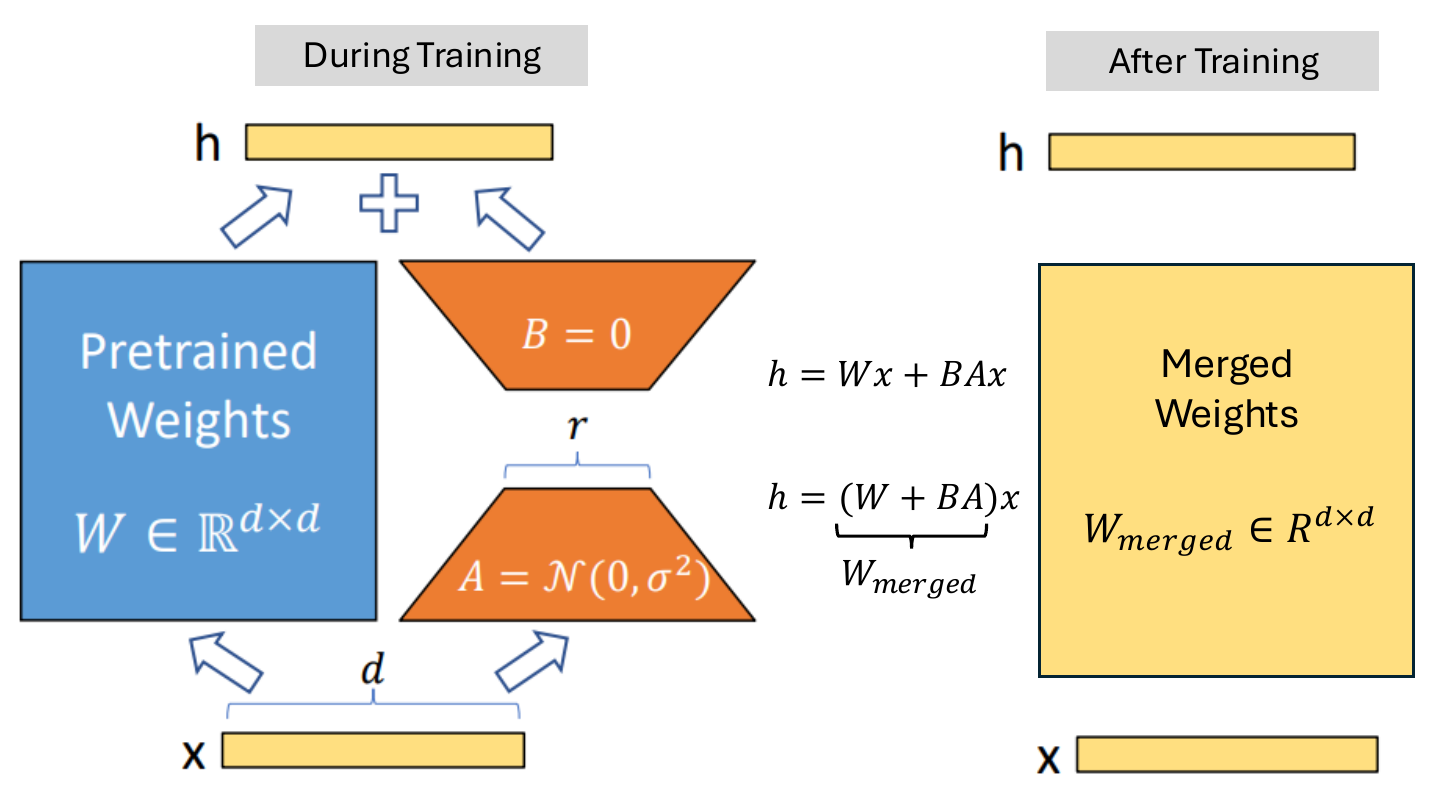}
\caption{LoRA training strategy \cite{hu2022lora}: During training (left), low-rank matrices $A$ and $B$ are introduced alongside frozen pre-trained weights. Only $A$ and $B$ are updated. After training (right), the modified weights are merged into $W_{\text{merged}}$ for inference, facilitating efficient deployment with modest memory and computational overheads.}
\label{fig:lora}
\end{figure}

To efficiently adapt large MLLM backbones to the PU staging task with limited data, we adopt LoRA~\cite{hu2022lora}. LoRA is a parameter-efficient fine-tuning method that inserts trainable low-rank matrices into selected layers of a frozen pre-trained model, avoiding the need to update all parameters. Specifically, weight updates are represented as a low-rank decomposition:

\begin{equation}
h = W x + \Delta W x = W x + B A x,
\end{equation}

where $W$ is the frozen pre-trained weight matrix, and $\Delta W = B A$ is the learned update composed of two low-rank matrices $A \in \mathbb{R}^{r \times d}$ and $B \in \mathbb{R}^{d \times r}$. During training, only $A$ and $B$ are updated (with $B = 0$ at initialization), and at inference, $W$ and $\Delta W$ are merged into a single matrix $W_{\text{merged}}$ for efficient deployment (Figure~\ref{fig:lora}).

In FT-ARM, LoRA is applied to the query and value projection layers of the LLM’s self-attention blocks, and optionally to corresponding layers in the vision encoder. We use a rank of $r = 8$, introducing only a small number of additional parameters relative to the full model size. This setup offers multiple advantages:  
(i) significantly lower memory and compute requirements,  
(ii) reduced risk of overfitting due to weight freezing, and  
(iii) retention of the general-domain knowledge encoded in the original MLLM.

\subsubsection{Practical Inference Adjustment}
\label{sec:inference-adjustment}

During preliminary experiments, we observed that  when optimized specifically for pressure ulcer stage classification, the fine-tuned multimodal LLM often predicted only the PU class label without providing explanatory reasoning, even when explicitly prompted. This behavior arose because the model was fine-tuned solely on stage classification labels, without exposure to corresonding pairs of textual rationales during training. Consequently, the fine-tuned model learned to prioritize accurate class prediction while neglecting the generation of associated descriptive rationale, which is not reinforced by the loss objective. To mitigate this and preserve interpretability, a post-finetuning inference strategy that decouples classification from rationale generation was introduced.

In this setup, the fine-tuned FT-LLM is first used to predict the most probable pressure ulcer stage given the wound image and task-specific prompt. The predicted stage label and prompt are then passed to the base (non-fine-tuned) version of the same MLLM, which has retained its broader reasoning capability. The base model generates the clinical rationale conditioned on both the prompt and the estimated class.

\begin{figure}[h]
    \centering
    \includegraphics[width=0.75\linewidth]{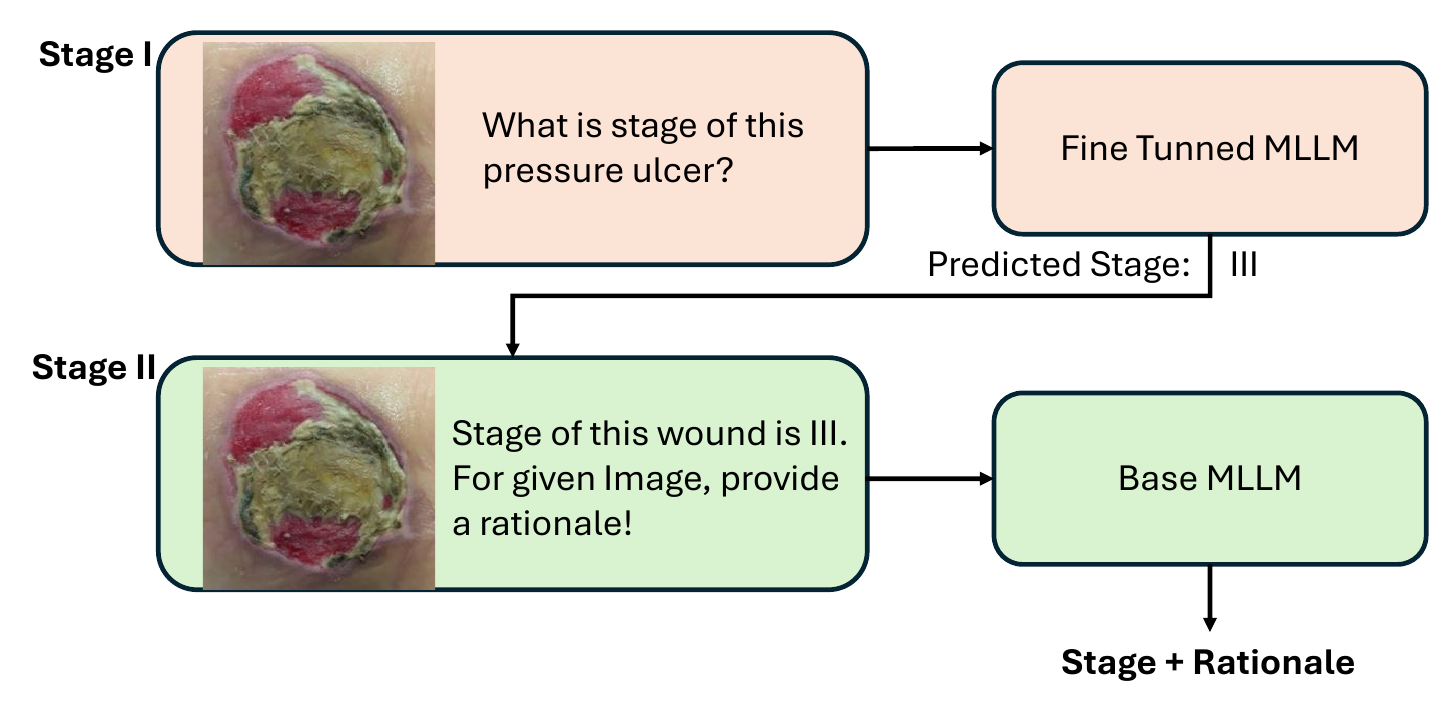}
    \caption{Illustration of FT-ARM's post-finetuning inference adjustment strategy. In \textbf{Stage I}, the fine-tuned MLLM receives the wound image and prompt \emph{``What is the stage of this pressure ulcer?''} and outputs a predicted stage (e.g., Stage III). In \textbf{Stage II}, the predicted stage and image are passed to the base MLLM with the reasoning prompt \emph{``The stage of this wound is III. For the given image, provide a rationale!''}, which produces both the stage and explanatory rationale. This decoupled inference workflow preserves interpretability without requiring additional reasoning-specific fine-tuning.}
    \label{fig:inference-adjustment}
\end{figure}

This hybrid inference design allows FT-ARM to combine the classification precision of the fine-tuned model with the reasoning depth of the base model, ensuring that each prediction remains both accurate and interpretable. Figure~\ref{fig:inference-adjustment} illustrates this two-step process, where the fine-tuned model outputs the stage classification and the base model subsequently generates the associated rationale.

\section{Evaluation and Experimental Results}
\label{sec:evaluation}
\subsection{Evaluation Metrics}

FT-ARM and baseline models were evaluated using two key metrics: overall classification \textbf{accuracy} and the \textbf{F\textsubscript{1}} score. 

\textbf{Accuracy} Accuracy measures the proportion of correct predictions across all classes and is defined as:
\begin{equation}
   \text{Accuracy} = \frac{TP + TN}{TP + TN + FP + FN}
\end{equation}

\textbf{$F_1$-score} provides a balanced measure of precision and recall, which is particularly important for clinical tasks with class imbalance and is defined as:
\begin{equation}
\text{Precision} = \frac{TP}{TP + FP}, \quad
\text{Recall} = \frac{TP}{TP + FN}
\end{equation}

\begin{equation}
\text{F}_1 = 2 \cdot \frac{\text{Precision} \cdot \text{Recall}}{\text{Precision} + \text{Recall}}
\end{equation}

\subsection{SOTA Baseline Models}
We compared FT-ARM to three categories of baseline models: (i) prior SOTA CNN-based methods, (ii) ViT-based methods, and (iii) recent MLLMs.

\textbf{CNN-Based Baselines:}
Based on their reported performance in prior work~\cite{ay2022deep,wang2024novel}, the following CNN architectures were included: MobileNetV2, VGG16, DenseNet121, ResNet152, EfficientNetV2-s, and ResNeXt50.  ResNeXt50 + wFPN~\cite{wang2024novel} was the best performer, achieving 81.5\% accuracy on the PIID dataset.

\textbf{ViT-Based Baseline:}
FT-ARM was also compared to the Swin Transformer-tiny~\cite{liu2021swin} as explored in~\cite{wang2024novel}, which achieved 75.5\% accuracy. Although it slightly underperformed the best CNN, it represents a strong transformer-based baseline.

\begin{figure}[h]
\centering
\includegraphics[width=0.95\linewidth]{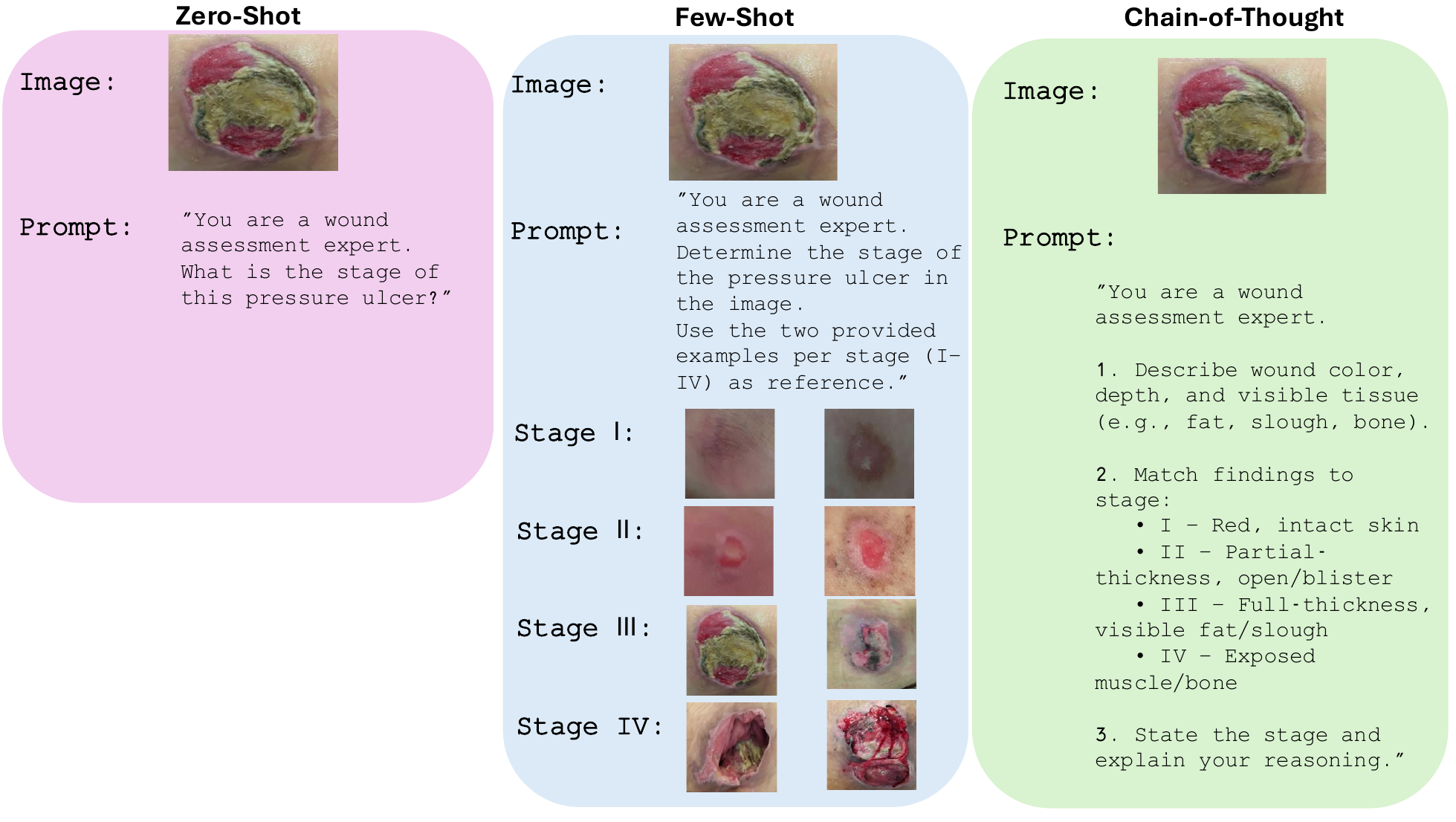}
\caption{Illustration of prompting strategies used to evaluate baseline MLLMs. \textbf{Left:} Zero-shot prompting with only a classification instruction. \textbf{Center:} Few-shot prompting provides two labeled examples per class (Stages I–IV) to guide the model. \textbf{Right:} Chain-of-Thought (CoT) prompting uses a structured reasoning prompt with step-by-step visual assessment and stage definitions.}
\label{fig:baseline-prompt-modes}
\end{figure}

\textbf{MLLM Baselines:} We evaluated five recent MLLMs using zero-shot, two-shot, and chain-of-thought (CoT) prompting modes. For Few-shot prompting, two labeled support examples per class (Stages I–IV) were included, yielding  a total of eight examples appended before the query image. For chain-of-thought prompting, structured instruction guided the model to first describe visual wound features, which were then matched to stage definitions. Finally, a diagnosis with rationale was output. Figure~\ref{fig:baseline-prompt-modes} illustrates representative examples of the various prompting approaches explored for the three evaluation modes.

These included:
\begin{itemize}
    \item \textbf{GPT-4o}~\cite{openai2024gpt4o} – OpenAI’s flagship MLLM capable of reasoning over image and text.
    \item \textbf{LLaMA 3.2}~\cite{meta2024llama3} – Meta’s open-source MLLM with strong performance on most tasks.
    \item \textbf{Pixtral-12B}~\cite{mistral2023mixtral} – A mixture-of-experts transformer designed for visual tasks.
    \item \textbf{Qwen 2.5}~\cite{alibaba2024qwen2} – Alibaba’s MLLM with support for CoT and image grounding.
    \item \textbf{DeepSeek VL2}~\cite{deepseek2024vl2} – A pioneer multimodal model, based on mixture-of-experts transformers, trained for open-ended visual QA.
\end{itemize}

These MLLMs were selected based on their performance on vision-language benchmarks and the availability of their source code for modification, which FT-ARM required.

\subsection{Training and Test Splits}
To ensure robustness, \textbf{5-fold cross-validation} was conducted on the full PIID dataset comprising all 1091 images. Each fold used a stratified split, allocating approximately 80\% for training and 20\% for testing per class.

\begin{table}[h]
\centering
\caption{Per-stage image count across cross-validation folds (training / test).}
\label{tab:folds}
\begin{tabular}{|c|c|c|c|c|c|c|}
\hline
\textbf{Stage} & \textbf{Total} & \textbf{Fold 1} & \textbf{Fold 2} & \textbf{Fold 3} & \textbf{Fold 4} & \textbf{Fold 5} \\
\hline
Stage I   & 230  & 184 / 46 & 184 / 46 & 184 / 46 & 184 / 46 & 184 / 46 \\
Stage II  & 313  & 251 / 62 & 251 / 62 & 251 / 62 & 251 / 62 & 248 / 65 \\
Stage III & 275  & 220 / 55 & 220 / 55 & 220 / 55 & 220 / 55 & 220 / 55 \\
Stage IV  & 273  & 219 / 54 & 219 / 54 & 219 / 54 & 219 / 54 & 216 / 57 \\
\hline
\textbf{Total} & \textbf{1091} & \textbf{874 / 217} & \textbf{874 / 217} & \textbf{874 / 217} & \textbf{874 / 217} & \textbf{868 / 223} \\
\hline
\end{tabular}
\end{table}

\subsection{FT-ARM Configuration and Fine-tuning}

FT-ARM was implemented by fine-tuning multiple, alternate, open-source MLLMs, each built on a transformer-based architecture with visual and language processing components. LoRA was employed to adapt these MLLMs efficiently~\cite{hu2022lora}. Each FT-ARM variant was trained for 20 epochs using the AdamW optimizer. The model was fine-tuned exclusively on stage classification, as the PIID dataset provides ground-truth labels only for PU stage and did not include annotated rationales. During inference, FT-ARM generates explanatory rationales in free-form style, but they were not supervised during training.

\begin{figure}[h]
\centering
\includegraphics[width=0.55\linewidth]{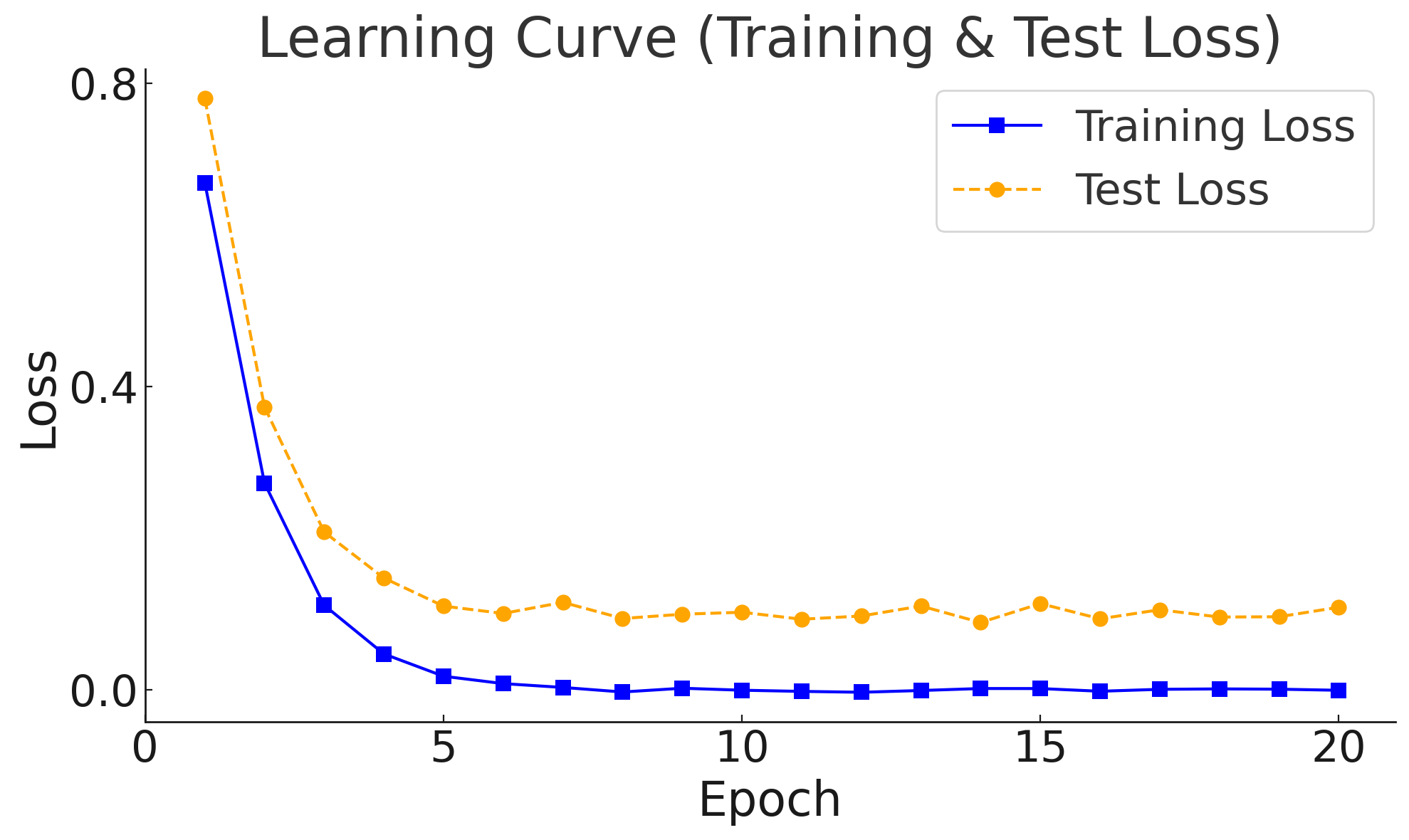}
\caption{Training and test loss curves for the LLaMA 3.2 90B backbone used in FT-ARM, fine-tuned on the PIID dataset over 20 epochs.}
\label{fig:train_curve}
\end{figure}

Figure~\ref{fig:train_curve} shows the training and test loss curves for the LLaMA 3.2 90B model fine-tuned on the PIID dataset as part of the FT-ARM framework. Over 20 epochs, the training loss steadily declined, while the test loss remained slightly higher but stable—indicating good generalization and no signs of overfitting. 

\begin{table}
\centering
\caption{FT-ARM training and inference hyperparameters.}
\label{tab:hyperparams}
\begin{tabular}{|l|l|}
\hline
\textbf{Hyperparameter} & \textbf{Value} \\
\hline
Image size & 224×224 \\
Visual encoder & ViT-base (frozen) \\
Language model & Transformer (\~10–13B) \\
LoRA rank & 8 \\
Trainable params & ~30M \\
Optimizer & AdamW, LR=2e-5, cosine decay \\
Batch size & 16 \\
Epochs & 20 \\
Reflection cycles & 1 \\
Decoding & Constrained decoding to known labels \\
\hline
\end{tabular}
\end{table}

At inference time, FT-ARM uses an \textbf{agentic reflection loop} for up to two iterations of self-correction, whereby it re-prompts itself to revise its answer if the critique expresses uncertainty about its output.

\subsection{Results}

\subsubsection{FT-ARM vs Baseline Models}

Table~\ref{tab:ftarm_vs_baseline} summarizes a comparison of the performance of FT-ARM and baseline models under various prompting and configurations. The FT-ARM variant that utilizes a fine-tuned LLaMA 90B backbone with an ARM achieves the highest overall accuracy of \textbf{85.2\%} and a $F_1$-score of \textbf{0.85} on the PIID dataset. Among baseline MLLMs, GPT-4o with chain-of-thought (CoT) prompting achieved the best result (accuracy 0.71, $F_1$ 0.78), followed by Qwen2-VL-72B and Pixtral-12B. However, all zero-shot and few-shot prompting strategies consistently significantly underperformed relative to FT-ARM, highlighting the importance of fine tuning. Crucially, ablative results highlight the individual and complementary contributions of FT-ARM’s components. Removing fine-tuning (ARM only) drops performance to 0.53 accuracy, while removing agentic reflection (FT only) yields 0.84—indicating that fine-tuning contributes more substantially on the PIID. Both components are necessary to achieve optimal performance, though their relative impact is task- and dataset-dependent and may differ for other clinical domains or datasets.

\begin{figure}[t]
\centering
\includegraphics[width=0.6\textwidth]{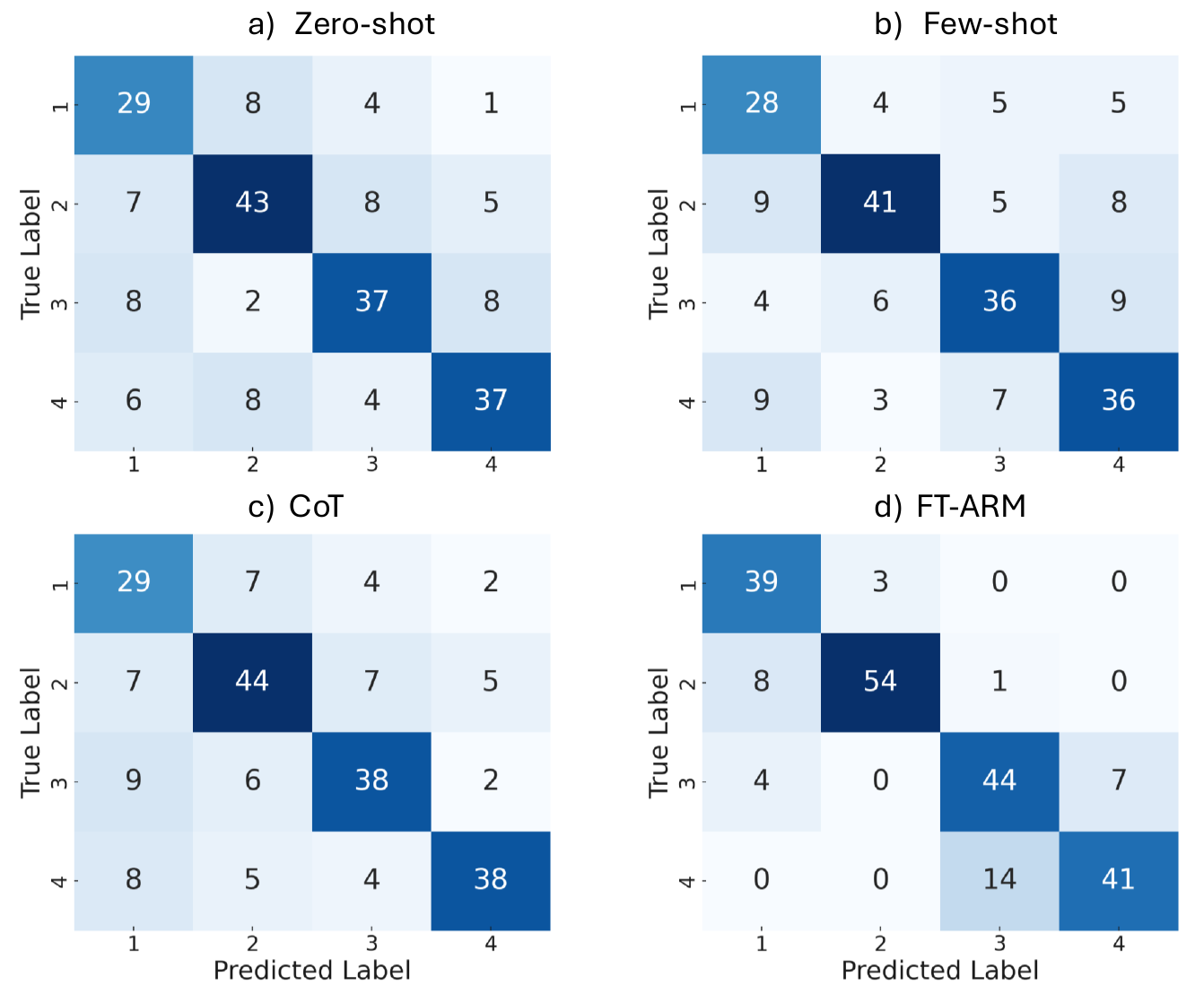}
\caption{Confusion matrices comparing predicted vs.\ true pressure ulcer stages on the test set for the proposed FT-ARM approach using different prompting strategies: Zero-shot, Few-shot, CoT, and FT-ARM. While baseline prompting strategies (best: GPT-4o) exhibit moderate classification ability, FT-ARM (best: LLaMA 3.2 90B) demonstrates the strongest alignment with ground truth, with a pronounced diagonal dominant pattern and minimal off-diagonal errors. This highlights FT-ARM's superior ability to accurately stage pressure ulcers. }
\label{fig:confusion}
\end{figure}

These values demonstrate improved robustness, especially in the most challenging and clinically severe cases. Figure~\ref{fig:confusion} further visualizes FT-ARM’s strong performance via confusion matrices—strong diagonal dominance in the prediction patterns and reduced mis-classification compared to prompting-based baselines. Compared to other prompting strategies (zero-shot, few-shot, and CoT), FT-ARM, which incorporates both fine-tuning and agentic reflection, and uses the same simple prompt as the zero-shot baseline method, exhibits a clearer dominant diagonal patterns and fewer off-diagonal errors. Notably, FT-ARM achieves higher classification consistency for Stages III and IV, which are often the most challenging to discriminate. This suggests enhanced reliability in discriminating more severe wound cases, which is critical in clinical practice.

\begin{table}[h]
\centering
\scriptsize 
\caption{FT-ARM vs. Baseline LLMs under Zero-shot, Few-shot, and Chain-of-Thought (CoT) prompting, and internal components: ARM (reflection only), FT (fine-tuning only), and FT-ARM (combined). NA = not applicable or not supported.}
\label{tab:ftarm_vs_baseline}
\begin{tabular}{|l|cc|cc|cc|cc|cc|cc|}
\hline
\textbf{Model} & \multicolumn{2}{c|}{\textbf{Zero-shot}} & \multicolumn{2}{c|}{\textbf{Few-shot}} & \multicolumn{2}{c|}{\textbf{CoT}} & \multicolumn{2}{c|}{\textbf{ARM}} & \multicolumn{2}{c|}{\textbf{FT}} & \multicolumn{2}{c|}{\textbf{FT-ARM}} \\
& ACC (\%) & F1 (\%) & ACC (\%) & F1 (\%) & ACC (\%) & F1 (\%) & ACC (\%) & F1 (\%) & ACC (\%) & F1 (\%) & ACC (\%) & F1 (\%) \\
\hline
GPT-4o & 68 ± 5 & 69 ± 6 & 66 ± 7 & 65 ± 8 & 70 ± 6 & 71 ± 6 & 78 ± 3 & 77 ± 4 & NA & NA & NA & NA \\
GPT-4o-mini & 62 ± 5 & 61 ± 4 & 67 ± 8 & 66 ± 6 & 63 ± 8 & 70 ± 7 & 70 ± 4 & 71 ± 5 & NA & NA & NA & NA \\
LLaMA 11B & 43 ± 9 & 41 ± 8 & NA & NA & 52 ± 8 & 53 ± 7 & 49 ± 6 & 48 ± 8 & 78 ± 4 & 79 ± 4 & 82 ± 3 & 82 ± 5 \\
LLaMA 90B & 35 ± 6 & 29 ± 8 & NA & NA & 29 ± 10 & 24 ± 8 & 53 ± 6 & 49 ± 5 & 84 ± 2 & 84 ± 3 & \textbf{85 ± 3} & \textbf{85 ± 4} \\
Pixtral-12B & 35 ± 8 & 23 ± 11 & NA & NA & 39 ± 7 & 36 ± 9 & 41 ± 8 & 42 ± 8 & 53 ± 7 & 55 ± 6 & 55 ± 7 & 51 ± 5 \\
Qwen2-VL-72B & 39 ± 6 & 37 ± 9 & 41 ± 8 & 40 ± 7 & 52 ± 6 & 51 ± 7 & 49 ± 7 & 50 ± 5 & 56 ± 5 & 57 ± 6 & 54 ± 4 & 49 ± 5 \\
DeepSeek VL2 & 30 ± 10 & 19 ± 11 & 33 ± 9 & 25 ± 10 & 35 ± 8 & 31 ± 10 & 36 ± 7 & 34 ± 9 & NA & NA & NA & NA \\
\hline
\end{tabular}
\end{table}

\begin{figure}[h]
\centering
\includegraphics[width=0.95\linewidth]{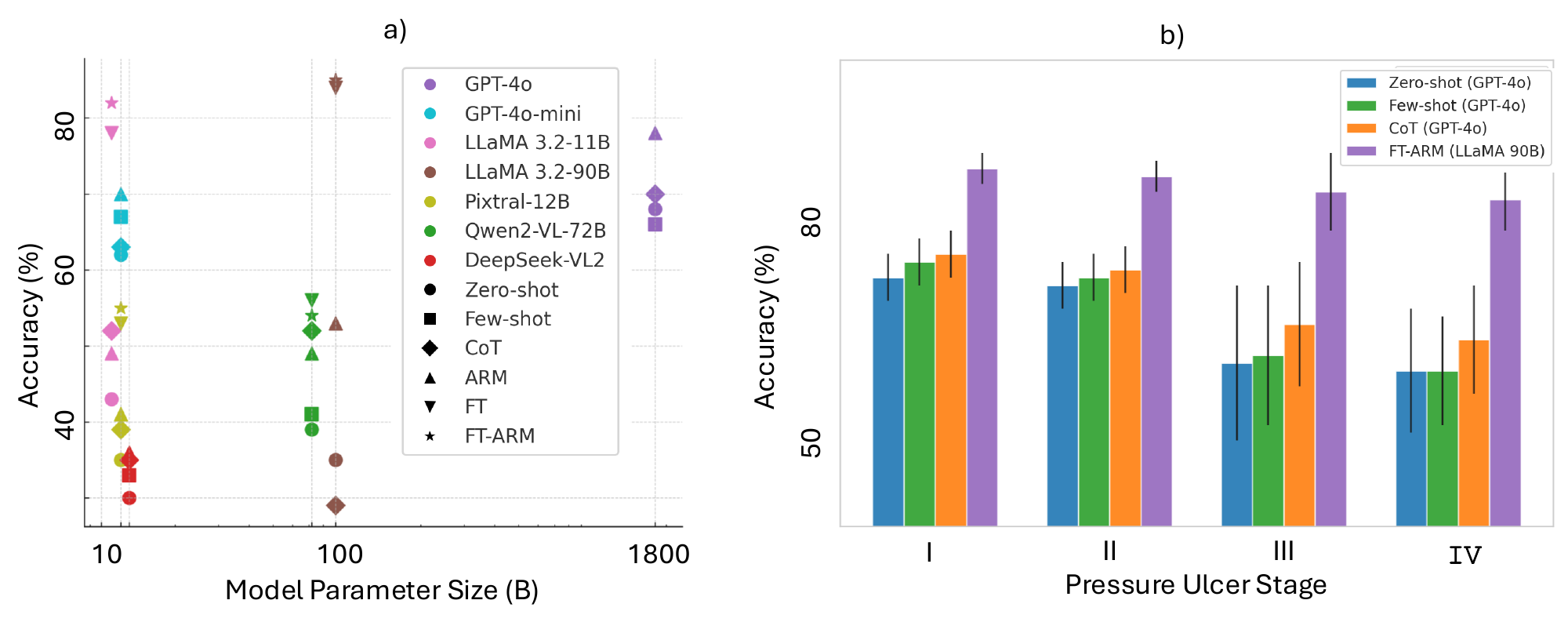}
\caption{\textbf{FT-ARM Performance vs. GPT-4o Prompting Baselines (LLaMA-90B vs. GPT-4o).}
(a) Accuracy vs. model size for multiple MLLMs and prompting strategies. FT-ARM using the smaller LLaMA 3.2–90B model as backbone MLLM, outperforms much larger models such as GPT-4o (estimated 1.8T parameters).
(b) Box plot of per-stage accuracy across 5 cross-validation folds. FT-ARM using Llama 3.2 90B as its backbone MLLM achieves superior and more consistent accuracy across all pressure ulcer stages (I–IV) compared to GPT-4o-based zero-shot, few-shot, and chain-of-thought prompting. The error bars capture performance variance, which is noticeably higher in worse-performing methods.}

\label{fig:ftarm-vs-baselines}
\end{figure}

Figure~\ref{fig:ftarm-vs-baselines} provides further insight into FT-ARM’s superior performance. In panel (a), despite using a smaller backbone MLLM (LLaMA with 90B parameters), FT-ARM outperforms much larger MLLMs such as GPT-4o,  demonstrating the effectiveness of targeted fine-tuning and agentic reflection for as pressure ulcer image classification.
Panel (b) compares the stage-wise accuracies across various prompting strategies using each of their best performing MLLM (GPT-4o for zero-/few-shot/CoT; LLaMA-90B for FT-ARM). FT-ARM consistently achieves the highest accuracy across all pressure ulcer stages. Notably, the performance superiority of FT-ARM is especially large for Stages III and IV, which are more challenging and frequently have clinically ambiguous cases. The narrower standard deviations of FT-ARM across folds reflects better generalization and stability, further supporting its robustness on the high-stakes wound staging task.

\subsection{Expert Clinician Validation of Results}
\label{sec:expert-validation}

To evaluate the clinical validity and interpretability of FT-ARM’s predictions, we conducted an expert review with a certified wound care nurse. A total of 84 wound images from the test set were initially considered. Since the purpose of this evaluation was to assess the quality and clinical soundness of the model’s generated rationales, only the subset of images for which FT-ARM correctly predicted the ground-truth stage labels and the nurse agreed with the ground truth were used in the analysis. Each selected image was accompanied by the model’s predicted stage and its automatically-generated explanatory rationale. The nurse, a board-certified podiatric physician and nurse practitioner with over thirty five years of expertise in vascular surgery and lower extremity wound care, examined each case to verify agreement with the ground-truth stage and subsequently rated the model’s explanatory rationale according to the categories defined in Table~\ref{tab:nurse-feedback-definitions}.

\begin{table}[h]
\centering
\caption{Definition of clinician feedback categories used to evaluate FT-ARM’s predictions and rationales for the selected cases.}

\label{tab:nurse-feedback-definitions}
\resizebox{0.9\linewidth}{!}{%
\begin{tabular}{p{2cm}p{12cm}}
\toprule
\textbf{Rating} & \textbf{Definition} \\
\midrule
\textbf{Good} & The generated rationale is clinically accurate, relevant, and clearly identifies wound characteristics consistent with the correct stage. \\
\textbf{Passable} & The rationale includes minor omissions, vague phrasing, or superficial clinical descriptions, yet remains mostly accurate and acceptable. \\
\textbf{Bad} & The rationale is clinically inconsistent, contains incorrect statements, or fails to describe the relevant wound features. \\
\bottomrule
\end{tabular}}
\end{table}

\begin{figure}[!ht]
\centering
\includegraphics[width=0.5\linewidth]{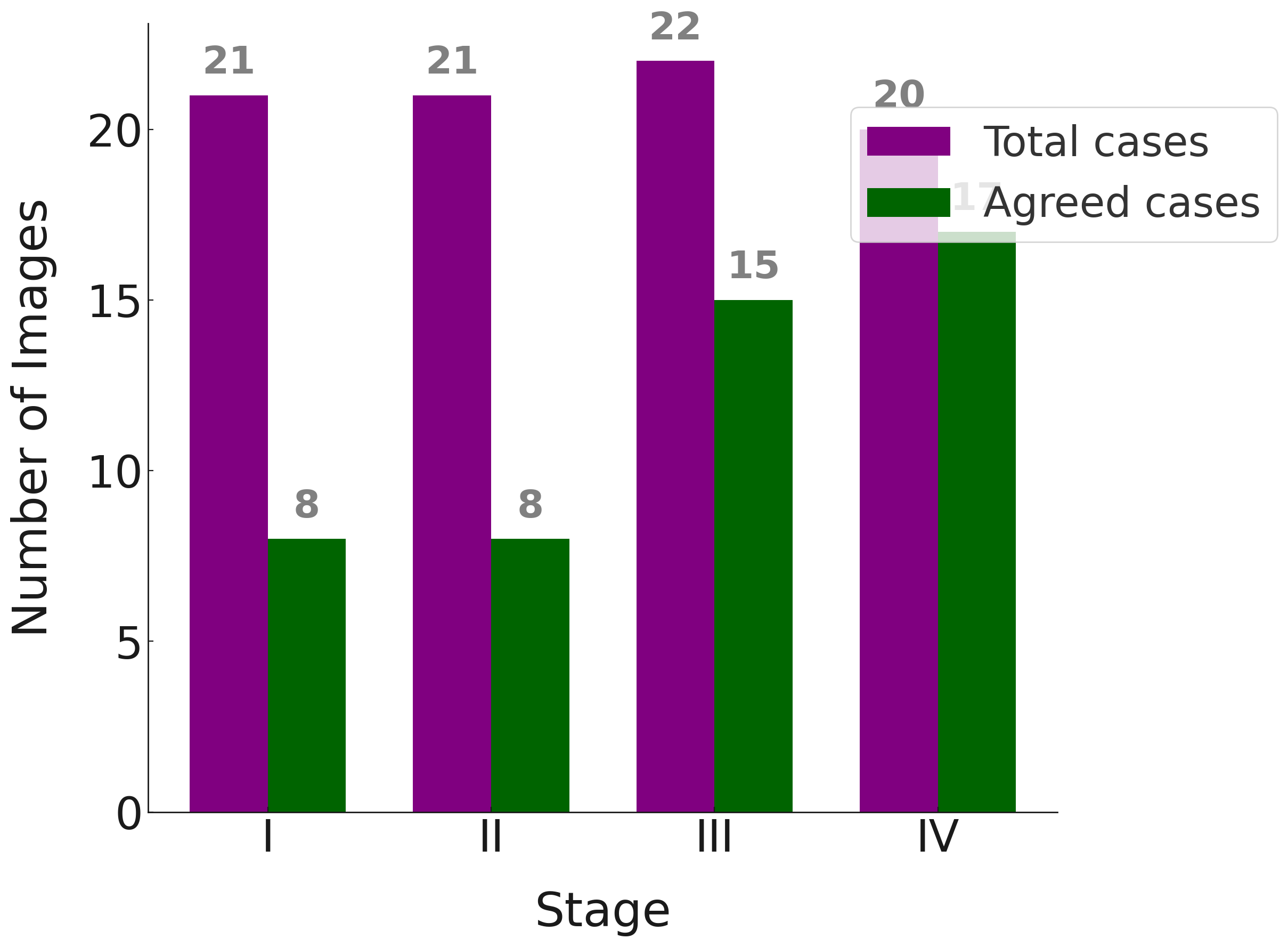}
\caption{Number of total and cases where there was agreement on the ground-truth PU labels per stage in the nurse validation set (n=84). Agreement rates were highest for advanced stages (III–IV), suggesting greater model reliability in severe wound cases.}
\label{fig:barchart-total-agreed}
\end{figure}

As shown in Figure~\ref{fig:barchart-total-agreed}, the nurse’s assessments matched the ground-truth labels in 57\% of the 84 test images (48/84). This subset where there was agreement on the ground-truth labels was used for rationale evaluation. In the remaining 36 cases (43\%), the nurse disagreed with the dataset labels, primarily due to (i) suspected annotation errors (wrong PU stage categorization), (ii) healing wounds that appeared visually improved yet retained their original clinical stage, and (iii) unstageable wounds with necrotic or slough tissue obscuring depth assessment. Because FT-ARM model was trained based on ground-truth labels in the dataset, these cases could not be used for explanation assessment, as they no longer represented valid targets relative to the model’s learning objective.. A detailed breakdown, including stage-wise statistics and representative examples of both agreement and disagreement, is provided in Appendix~\ref{sec:appendix-disagree}.

\begin{figure}[!ht]
\centering
\includegraphics[width=0.7\linewidth]{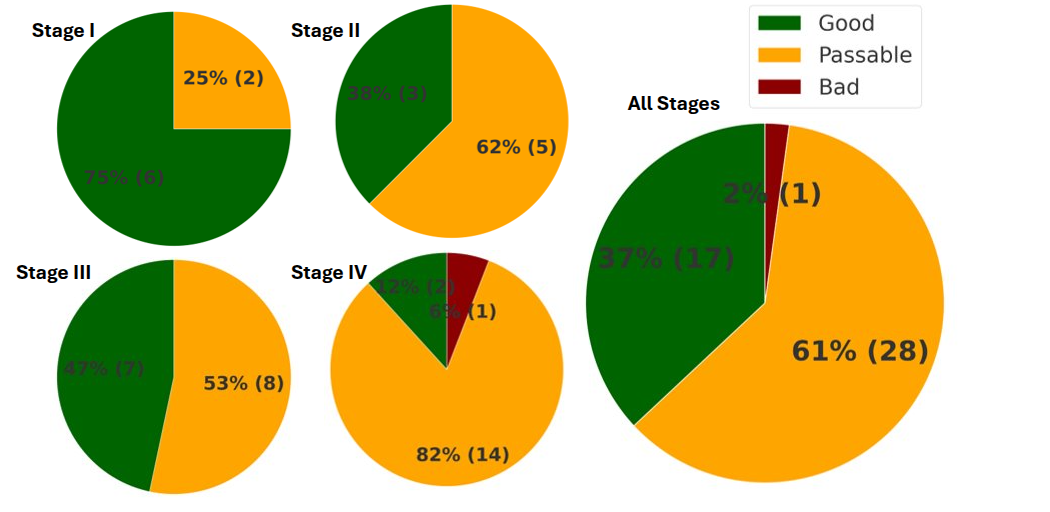}
\caption{Distribution of nurse feedback on the quality of the rationale generated by FT-ARM for the cases where the nurse agreed with the datasets ground truth PU label (n=48). The large pie chart on the right summarizes ratings across all stages, while the four smaller charts (top left to bottom right) correspond to Stage~I, Stage~II, Stage~III, and Stage~IV, respectively. The majority of rationales were rated as “Passable” or “Good,” indicating overall acceptable clinical reasoning with only one “Bad” case observed.}

\label{fig:piechart-feedback}
\end{figure}

Of the 48 cases with agreement on the PU ground truth labels, the nurse further evaluated the quality of the rationale generated by FT-ARM. As illustrated in Figure~\ref{fig:piechart-feedback}, 17 (35\%) were rated as “Good,” 28 (58\%) as “Passable,” and 3 (6\%) as “Bad.” This distribution indicates that while the model’s reasoning is generally clinically coherent, some rationales lack sufficient descriptive depth or precision in identifying subtle wound features. The only “Bad” case was due to incorrect clinical reasoning and  incomplete or ambiguous phrasing.

Overall, the expert validation confirmed that FT-ARM generates clinically valid stage predictions supported by mostly accurate rationales. Higher agreement in advanced stages highlights the model’s ability to capture deeper structural cues of wound severity such as tissue loss, while disagreement analysis emphasizes the importance of refining dataset annotations and incorporating more nuanced clinical context in future training.

\subsection{Interpretability}

\begin{figure}[t]
\centering
\includegraphics[width=0.9\textwidth]{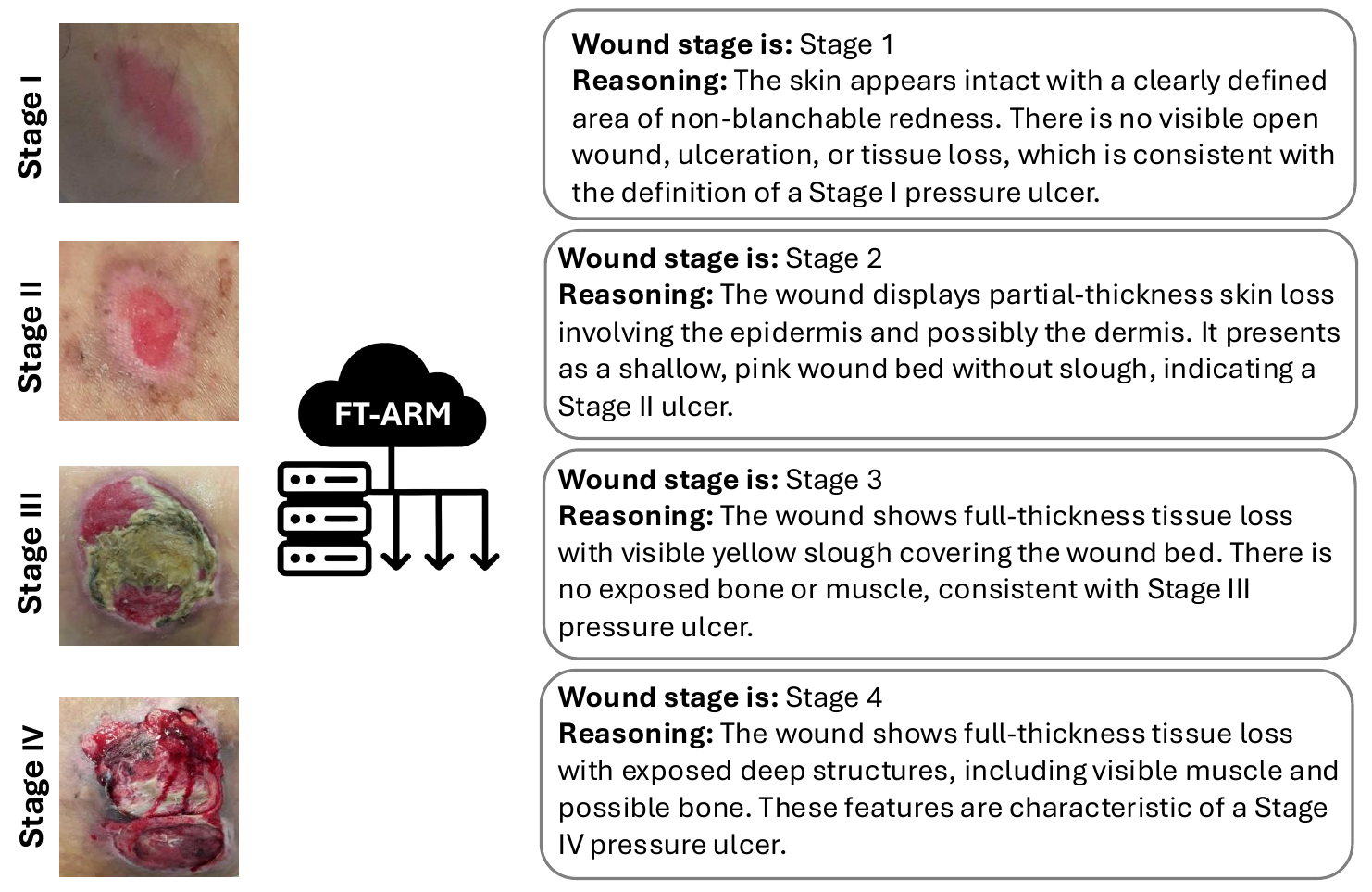}
\caption{Illustrative examples of FT-ARM’s stage prediction and reasoning for all four pressure ulcer stages. For each stage (I–IV), FT-ARM provides both a pressure ulcer stage prediction as well as corresponding rationale that highlights important clinical descriptors such as non-blanchable redness, tissue depth, presence of slough, and necrotic exposure. These example outputs demonstrate how FT-ARM generates pressure-ulcer stage-appropriate, human-understandable, natural language explanations that align with clinical assessment criteria, fostering trust in FT-ARM's decisions.}
\label{fig:interpret}
\end{figure}

A key advantage of FT-ARM is its ability to generate clinically grounded rationale corresponding to each pressure ulcer stage classification.  Unlike unmodified, baseline LLMs that often produce vague or error-prone explanations and hallucinations (making up wrong information), FT-ARM generates specific, stage-appropriate descriptions that mirror clinical reasoning. These rationales highlight relevant wound features—such as tissue integrity, presence of slough, visibility of underlying structures, and color changes—that are critical for pressure ulcer staging. 
Figure~\ref{fig:interpret} shows representative examples of rationale generated by FT-ARM for all four pressure ulcer stages. For each case, the model correctly identifies the stage and also generates concise rationale in natural language that highlights clinically relevant features—such as non-blanchable erythema, partial or full-thickness tissue loss, slough, and necrosis. These explanations are produced by the FT-ARM Generator LLM and align closely with expert definitions and staging criteria. Such interpretability fosters clinician trust in the model’s outputs and supports effective decision-making in clinical settings.

\section{Discussion}
\label{sec:discussion}


\subsection*{Performance and Reliability}

\textbf{FT-ARM surpasses prior CNN-based methods and approaches that involve prompting of unmodified SOTA LLMs} such as GPT-4o and Qwen 2.5, achieving an accuracy of 85.2\% on the PIID dataset. This performance underscores the importance of domain-specific adaptations.  Generalist models frequently struggled with discriminating fine-grained images (similar images that belong to different classes). Such images typically have subtle distinctions between pressure ulcer stages that are crucial for accurate staging. In contrast, FT-ARM was able to correctly identify subtle morphological features—e.g., discriminating Stages II and III ulcers—that often confound non-specialized models. This improvement can be attributed to both targeted fine-tuning via LoRA and the incorporation of reflective reasoning (ARM), which provided a mechanism for self-correction.

\textbf{Fine-tuning enables domain-specific visual and linguistic adaptations.} The FT component of FT-ARM trains the base LLM on expert-annotated wound images and clinical prompts, enabling it to specialize on the task by identifying relevant features such as slough, tissue depth, necrosis, and skin erythema. This adaptation improves both classification performance and the clinical alignment of model-generated rationale. As shown in Table~\ref{tab:ftarm_vs_baseline}, fine-tuning substantially boosts accuracy: for example, fine-tuning (FT) increased the accuracy of LLaMA 11B from 49\% (ARM only) to 78\%, demonstrating the critical role of task-specific adaptation. Unlike generic prompting-only LLM approaches, fine-tuned models internalize medically significant cues and produce outputs that are more consistent with expert reasoning.

\textbf{FT-ARM's agentic reflection loop contributed meaningfully to prediction robustness}. In multiple cases, FT-ARM revised its initial classification upon reflection (see Figure~\ref{fig:agentic-example} for a representative example) mirroring a clinician’s cognitive process of reviewing ambiguous evidence. In the illustrated case, FT-ARM initially predicted Stage IV due to visible slough and depth, but after a critique, this classification was revised correctly to Stage III based on the absence of exposed bone or muscle. While the mechanism was not universally corrective—occasionally reinforcing initial misjudgments when visual cues were ambiguous—it generally improved consistency and reduced errors.

\subsection*{Interpretability and Clinical Trust}

A central design goal of FT-ARM was to improve transparency in clinical AI systems by generating interpretable, medically relevant justifications for its predictions. Although a formal expert review was not conducted for this study, qualitative inspection confirmed that most rationales generated by FT-ARM referenced appropriate clinical cues—such as tissue type, wound depth, or the presence of slough—and utilized medically meaningful and relevant language. These explanatory outputs help address concerns about the “black-box” nature of deep learning models, and may improve trust and usability in real-world clinical settings. Future work will involve systematic validation of these rationales by wound care professionals, in line with prior clinical evaluation methodologies~\cite{palawat2024evaluation}.  

Textual rationale often cited pertinent visual indicators, such as skin integrity, wound depth, or tissue characteristics, and used appropriate medical terminology. Unlike many of the baseline unmodified MLLMs, rationale generated by FT-ARM avoided vague or generic descriptions and offered interpretable insights into its decision-making process. Additionally, FT-ARM tended to generate conservative outputs for ambiguous cases. This included sometimes overestimating a Stage II wound as Stage III (increased severity), but rarely the reverse. While not explicitly optimized for safety, intuitively, this behavior may align with clinical priorities, where caution by mistakingly adjudging a case to be more severe than its true severity, is more acceptable than the reverse.

The clinical validation study confirmed that when FT-ARM’s predictions matched the ground truth, the nurse generally judged its explanatory outputs as interpretable and clinically sound. Of the cases where the nurse agreed with the dataset's ground truth PU label (n=48), 35\% of rationales were rated as “Good,” 58\% as “Passable,” and only 2\% (one case) as “Bad”. Overall, the nurse noted that in most cases FT-ARM referenced appropriate wound characteristics—such as tissue depth, slough, and skin integrity—using medically meaningful terminology. These findings suggest that FT-ARM’s rationales provide clinically aligned explanations that can provide decision support for wound assessment.

\subsection*{Limitations}

Despite promising results, FT-ARM has several limitations:

\begin{itemize}
\item \textbf{Generalizability:} FT-ARM was trained and evaluated on a single dataset (PIID) consisting of 1,091 images. While our results are encouraging, generalization to other settings—datasets, hospitals, imaging equipment, and skin tones—remains unevaluated. Further validation on more diverse and fully representative datasets is required.

\item \textbf{Label Ambiguity:} 
Even experienced clinicians may occasionally disagree on the exact stage of a wound, reflecting the inherent subjectivity of pressure ulcer classification. In our evaluation, the wound care nurse agreed with the dataset’s ground-truth labels for 48 out of 84 test images and disagreed on the remaining 36. These disagreements highlight that the dataset had noisy labels with annotation inconsistency. A detailed analysis of these disagreements and representative examples are provided in Appendix~\ref{sec:appendix-disagree}. This limitation underscores the importance of rigorous data curation and labeling, and suggests that future datasets could benefit from multi-expert consensus labeling or uncertainty-aware annotation protocols to improve label reliability.

\item \textbf{Computational Overhead:} FT-ARM's backbone is a 90B-parameter LLM. While fine-tuning with LoRA made training efficient, inference remains resource-intensive. The reflection loop approximately doubles or triples inference time per image depending on the number of iterations. For example, on an Nvidia V100 GPU, inference  with the LLaMA 90B model without ARM takes approximately 2.35 seconds per image. With two iterations of the reflection loop, FT-ARM’s inference time increased to about 9.23 seconds. As shown in Table~\ref{tab:h100_inference_timing}, which reports both per-iteration and cumulative inference times on an Nvidia H100 GPU, the first iteration takes about 2.95 seconds, and each additional iteration adds roughly 2.85–2.90 seconds. For example, with three iterations, FT-ARM's inference time totals approximately 8.69 seconds. This may hinder real-time deployment in low-resource settings.

\begin{table}[h]
\label{tab:h100_inference_timing}
\centering
\caption{Inference time per reflection iteration using LLaMA 90B on an Nvidia H100 GPU.}
\begin{tabular}{lcc}
\toprule
\textbf{Iteration} & \textbf{Time (s) per Iteration} & \textbf{Cumulative Time (s)} \\
\midrule
1 & 2.95 $\pm$ 1.21 & 2.95 \\
2 & 2.85 $\pm$ 0.19 & 5.80 \\
3 & 2.89 $\pm$ 0.19 & 8.69 \\
4 & 2.89 $\pm$ 0.20 & 11.58 \\
5 & 2.90 $\pm$ 0.20 & 14.48 \\
\bottomrule
\end{tabular}
\label{tab:h100_inference_timing}
\end{table}

    \item \textbf{Variability of explanations and hallucination:} Although FT-ARM’s rationales were mostly accurate and informative, occasional hallucinations or omissions occurred. Some rationale generated with confident prose referenced features that were not present in the image. This necessitates clinician oversight of FT-ARM and incorporation of guardrails. While rare, such errors underscore the need for human verification and further validation of FT-ARM.
    
    \item \textbf{Scope of Task:} FT-ARM is currently limited to staging pressure ulcers from a single image. It does not yet support other important, desirable wound image analyses tasks such as lesion detection, holistic wound evaluation (e.g., infection status, size, patient context), or longitudinal monitoring. Integration of FT-ARM with diagnostic systems that perform a broad range of assessments could be considered to enhance clinical utility.
\end{itemize}

\section{Conclusion and Future Works}
\label{sec:conclusion}
\subsection*{Conclusion}

This paper proposed~\textbf{FT-ARM}: a Fine-Tuned, Agentic Reflection Multimodal model for classifying the severity of a pressure ulcer in a smartphone image with associated rationale in textual, clinically-valid, in natural language. By fine-tuning an MLLM with LoRA and incorporating a self-reflective reasoning ARM loop, FT-ARM achieved a classification accuracy of 85.2\% on PIID, the only publicly available pressure ulcer image dataset with expert-provided ground-truth labels. In rigorous evaluations, FT-ARM outperformed all baselines that included conventional CNNs and SOTA unmodified, generalist MLLMs using various prompting strategies.
Importantly, unlike prior machine learning approaches whose results were obtained from static offline test sets, FT-ARM is explicitly designed for and evaluated under live inference conditions. This ensures that the reported accuracy more accurately reflects real-world, live deployment scenarios, where input variability and ambiguity often degrades performance observed in offline evaluations. For example, in chest X-ray pneumonia detection, Zech et al.\ \cite{zech2018variable} reported that a CNN with an initial AUC of 0.931 showed substantially lower performance (AUC 0.815) on an external datasets containing images from a different hospital. This illustrates how in-distribution test set performance can often overstate model performance in the real world.

Beyond raw performance, FT-ARM's human-understandable, clinically-valid rationale  address a critical need for transparency in medical AI to enhance clinician trust. Notably, FT-ARM's  agentic reflection mechanism improved prediction reliability by enabling the model to reassess ambiguous cases in a fashion analogous to human clinical reasoning.
FT-ARM thus demonstrates how domain-specific fine tuning and reflective reasoning can enhance both the accuracy and interpretability of AI systems in healthcare. Its ability to “show its work” makes it more trustworthy and usable in settings such as wound care where decisions  must be both correct as well as explainable.

\subsection*{Future Work}

Building on this foundation, our future work will focus on improving generalizability, usability, and clinical integration. First, we plan to validate FT-ARM on additional external datasets collected in  diverse care settings, including images showing less common categories such as unstageable or deep tissue injuries. This will assess the FT-ARM’s robustness as well as adaptability beyond the PIID dataset. We also aim to extend our proposed FT-ARM framework to related tasks such as diabetic foot ulcer classification, burn depth estimation, and broader dermatologic assessments—domains where explainability is equally important. Additionally, we plan to explore extensions to the classification scheme itself, incorporating categories such as “healed” or incorporating wound size estimation.

Technically, we plan to  refine the agentic reflection loop by incorporating uncertainty estimation and user-triggered re-evaluation. We envision a more interactive version of the model, capable of asking clarifying questions (e.g., regarding image quality) or incorporating clinician feedback mid-inference. We also plan to explore the integration of clinical context such as nursing clinical notes or patient history into FT-ARM's input to further enhance its reasoning and staging accuracy.
Finally, we plan to conduct human-in-the-loop evaluations with wound care professionals and evaluate FT-ARM's utility in real clinical workflows. These studies will assess not only the model’s predictive performance in pressure ulcer staging but also the practical utility of the rationale it generates, and their influence on clinician decision-making, workflow efficiency, accuracy as part of auto-generated documentation, and ultimately, long term outcomes.

In summary, FT-ARM represents a promising step toward interpretable and trustworthy medical AI for wound care. To continue its development and validation, we plan to deploy it as a decision support tool to enhance the quality of care provided by clinicians with limited wound expertise.

\section*{Acknowledgments}
This work is supported by the National Institutes of Health (NIH) under grant 1R01EB031910-01A1. 



\bibliographystyle{ACM-Reference-Format}
\bibliography{main}

\clearpage
\begin{appendices}
\appendixpage

\section{Representative Agreement Examples}
\label{sec:appendix-agree}

As part of the expert clinician validation (Section~\ref{sec:expert-validation}), a certified wound care nurse reviewed 84 wound images from the test set to assess both the correctness and interpretability of FT-ARM’s predictions. Each image was accompanied by the model’s predicted stage and its automatically generated explanatory rationale, which the nurse evaluated using the rating criteria defined in Table~\ref{tab:nurse-feedback-definitions} and summarized in Figure~\ref{fig:piechart-feedback}.

Of the 84 reviewed cases, the nurse’s assessments agreed with the dataset's ground-truth labels in 48 instances (57\%), corresponding to the subset where FT-ARM’s predictions were also correct. These cases where there was label agreement were used for rationale quality evaluation, yielding 35\% rated as “Good,” 58\% as “Passable,” and 2\% (one case) as “Bad.” The single “Bad” case resulted from incorrect clinical reasoning rather than ambiguous or incomplete phrasing.

\begin{figure}[!ht]
\centering
\includegraphics[width=.9\linewidth]{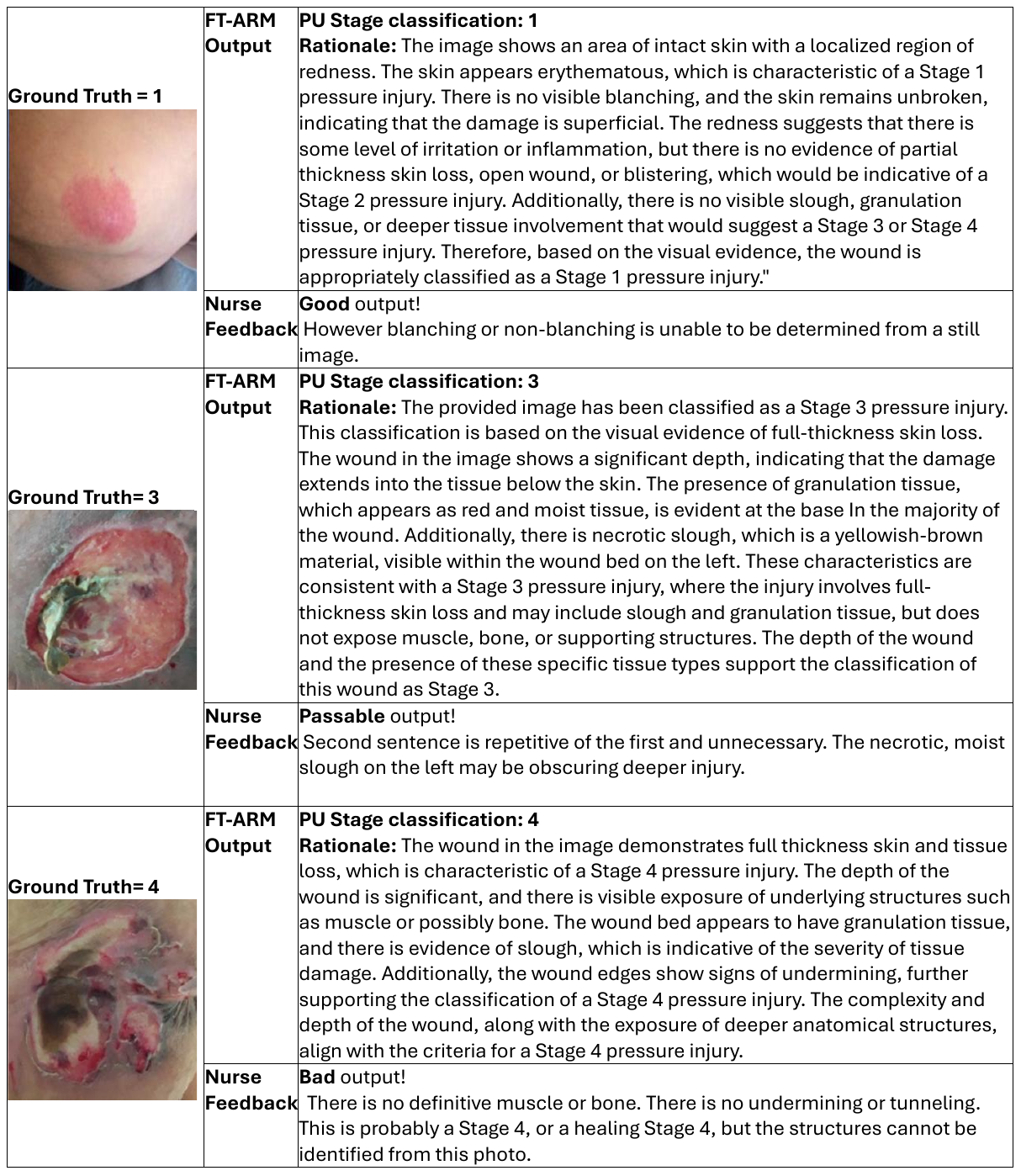}
\caption{Representative examples of cases where the nurse agreed with the ground truth labels in the dataset across various PU stages. The top example (\textbf{Stage I}) was rated as “Good,” demonstrating accurate recognition of intact skin with superficial erythema. The middle example (\textbf{Stage III}) was rated as “Passable,” showing generally correct reasoning with minor redundancy and possible depth ambiguity. The bottom example (\textbf{Stage IV}) was rated as “Bad,” where the rationale contained inaccurate clinical interpretation despite correct stage prediction. These examples illustrate how FT-ARM’s rationales vary in clinical precision and depth even when stage predictions are correct.}
\label{fig:agree-examples}
\end{figure}

\section{Representative Disagreement Analysis}
\label{sec:appendix-disagree}

\begin{figure}[!ht]
\centering
\includegraphics[width=.85\linewidth]{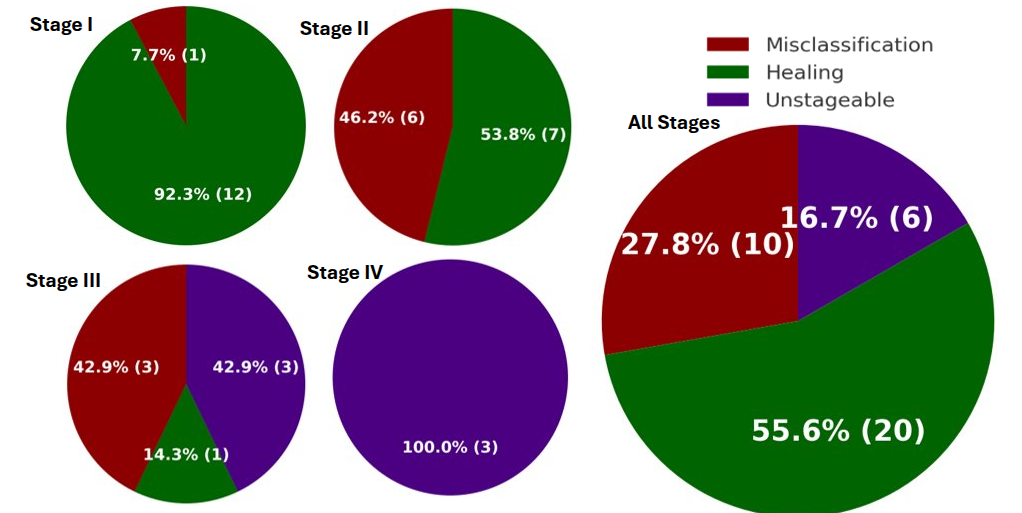}
\caption{Distribution of cases where the nurse disagreed with the ground truth PU stages in the dataset across various PU stages. Healing and unstageable wounds were the most common causes of disagreement, particularly in advanced stages (III–IV).}
\label{fig:pie-mis-heal-unstage}
\end{figure}

Not all test cases achieved full agreement between the dataset labels, FT-ARM predictions, and the nurse’s clinical interpretation. Discrepancies primarily fell into three categories: (i) \textbf{mis-categorization}, where dataset labels differed from the nurse’s judgment; (ii) \textbf{healing wounds}, which visually appeared to have improved but retained their original stage labels; and (iii) \textbf{unstageable cases}, where necrotic or slough tissue obscured depth assessment. A summary of these disagreement types is provided in Table~\ref{tab:disagreement-types}, and their distribution across wound stages is shown in Figure~\ref{fig:pie-mis-heal-unstage}.

\begin{table}[h]
\centering
\caption{Summary of categories where disagreement with the nurse's evaluation was observed.}
\label{tab:disagreement-types}
\resizebox{0.9\linewidth}{!}{%
\begin{tabular}{p{3cm}p{11cm}}
\toprule
\textbf{Disagreement Type} & \textbf{Definition and Clinical Description} \\
\midrule
\textbf{Misclassification} & The dataset stage label differs from the nurse’s expert judgment (e.g., annotated as Stage~II but exhibiting deeper tissue loss consistent with Stage~III). Represents labeling inconsistency rather than model error. \\
\textbf{Healing (Reverse Staging)} & Wounds showing epithelial ingrowth, hypopigmented margins, or contraction indicative of healing. Such wounds may appear visually shallower but remain clinically at their original stage. \\
\textbf{Unstageable} & Necrotic or slough tissue obscures the wound base, preventing assessment of depth or exposure of deeper structures. Cannot be reliably staged until debridement. \\
\bottomrule
\end{tabular}}
\end{table}

\begin{figure}[!ht]
\centering
\includegraphics[width=.90\linewidth]{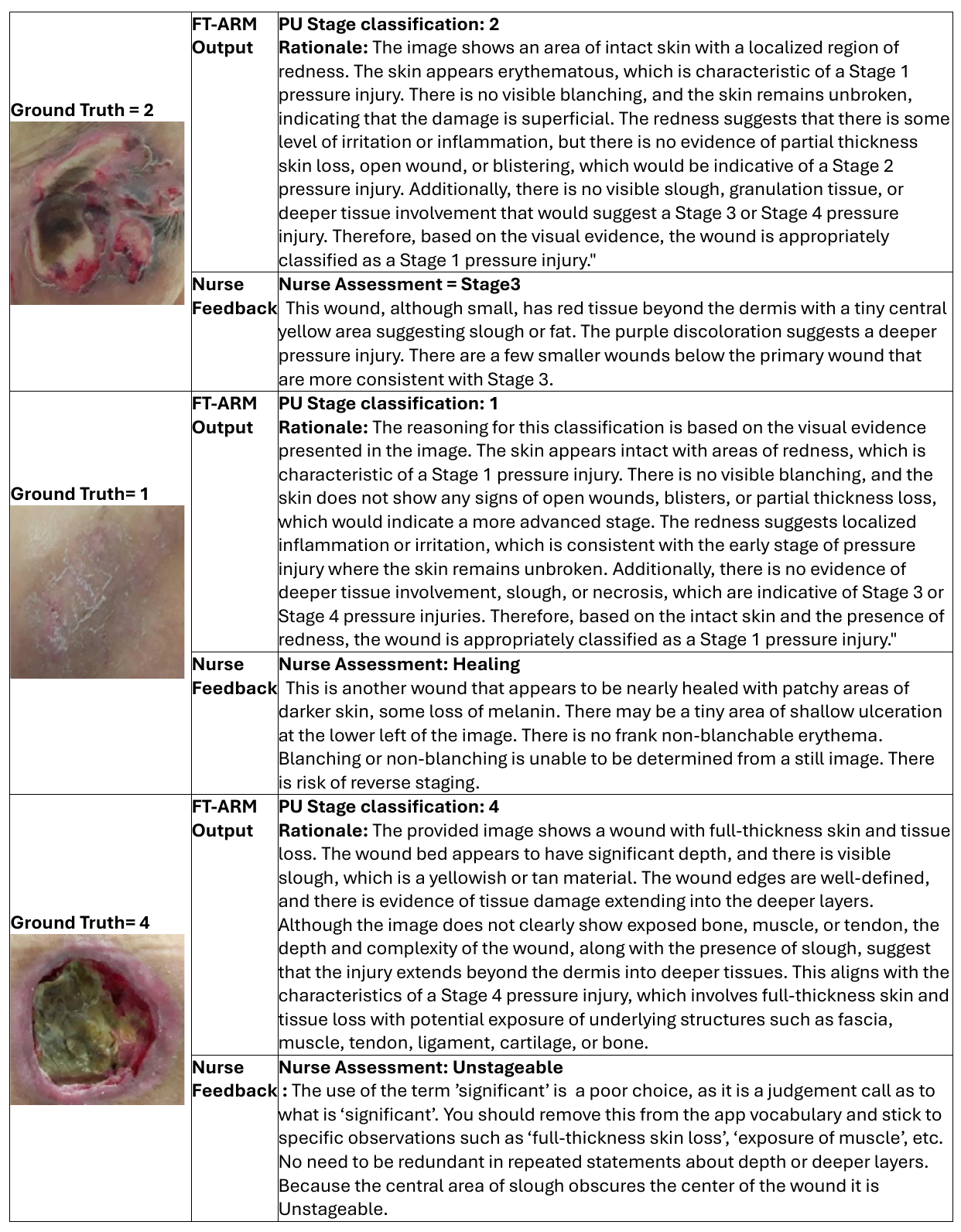}
\caption{Representative examples of cases where the nurse disagreed with the ground truth PU labels in the dataset: (top) dataset mis-categorization, (middle) healing wound (reverse staging), and (bottom) unstageable case. Each example includes FT-ARM’s rationale and the nurse’s clinical feedback highlighting the reason for disagreement.}
\label{fig:disagree-examples}
\end{figure}

Figure~\ref{fig:pie-mis-heal-unstage} shows that healing wounds accounted for over half of all disagreement cases, followed by mis-categorization and then the unstageable categories. Stage~III and IV wounds were more often labeled as unstageable or healing, while Stage~II had the highest proportion of mis-categorized labels. These disagreements reflect inherent labeling discrepancies rather than model errors. Representative examples of each of these categories are shown in Figure~\ref{fig:disagree-examples}.

\end{appendices}

\end{document}